\documentclass[runningheads]{llncs}

\usepackage[table]{xcolor}
 
\usepackage{eccv}

\usepackage{eccvabbrv}

\usepackage{multirow}
\usepackage{makecell}
\usepackage{graphicx}
\usepackage{booktabs}

\usepackage{pifont}%
\usepackage{colortbl}
\usepackage{amsfonts}

\definecolor{cvprblue}{rgb}{0.21,0.49,0.74}
\usepackage[breaklinks,colorlinks,allcolors=cvprblue]{hyperref}

\usepackage{orcidlink}

\newcommand{\yesmark}{\textcolor{green}{\ding{51}}}%
\newcommand{\nomark}{\textcolor{red}{\ding{55}}}%
\newcommand{\methodname}[0]{REALM\xspace}

\definecolor{rowgray}{gray}{0.95} 
\definecolor{rowblue}{HTML}{9FC0F2} %

\newcommand{\colortab}[0]{\cellcolor{rowblue!40}}
\newcommand{\colorrow}{\rowcolor{rowblue!40}}

\newcommand{\orange}[1]{{\color{orange}#1}}
\newcommand{\green}[1]{{\color{green!70!black}#1}}

\newcommand{\tb}[1]{\textbf{#1}}

\newcommand{\ul}[1]{\underline{#1}}

\renewcommand{\sectionautorefname}{Sec.}

\usepackage[absolute]{textpos}

\begin{document}

\definecolor{somegray}{rgb}{0.5, 0.5, 0.5}
\newcommand{\darkgrayed}[1]{\textcolor{somegray}{#1}}
\begin{textblock}{8}(4, 0.7)
\begin{center}
\darkgrayed{This paper has been accepted for publication at the \\
European Conference on Computer Vision (ECCV), Malmö, SE, 2026. \\
\copyright\ Springer Nature Switzerland AG 2026}
\end{center}
\end{textblock}

\title{REALM: An RGB- and Event-Aligned Latent Manifold for Cross-Modal Perception}
\titlerunning{An RGB- and Event-Aligned Latent Manifold for Cross-Modal Perception}

\author{
Vincenzo Polizzi$^{1}$ \and David B.\ Lindell$^{2}$ \and
Jonathan Kelly$^{1}$
\\
{\tt\small \{vincenzo.polizzi,jonathan.kelly\}@robotics.utias.utoronto.ca} \\
{\tt\small lindell@cs.toronto.edu}
}
\authorrunning{V.~Polizzi et al.}

\institute{$^{1}$University of Toronto, Robotics Institute \\
$^{2}$University of Toronto, Department of Computer Science}

\maketitle

\begin{abstract}
Event cameras provide several unique advantages over standard frame-based sensors, including high temporal resolution, low latency, and robustness to extreme lighting. 
However, existing learning-based approaches for event processing are typically confined to narrow, task-specific silos and lack the ability to generalize across modalities.
We address this gap with \methodname, a cross-modal framework that learns an \textbf{R}GB- and \textbf{E}vent-\textbf{A}ligned \textbf{L}atent \textbf{M}anifold by projecting event representations into the pretrained latent space of RGB foundation models.
Instead of task-specific training, we leverage low-rank adaptation (LoRA) to bridge the modality gap, effectively unlocking the geometric and semantic priors of frozen RGB backbones for asynchronous event streams.
We demonstrate that \methodname effectively maps events into the ViT-based foundation latent space. 
Our method performs downstream tasks, such as depth estimation and semantic segmentation, by simply transferring linear heads trained on the RGB teacher.
Most significantly, \methodname enables the direct, zero-shot application of complex, frozen image-trained decoders, such as MASt3R, to raw event data. We demonstrate state-of-the-art performance in wide-baseline feature matching, significantly outperforming specialized architectures. 
Code and models are available at \href{https://papers.starslab.ca/realm}{https://papers.starslab.ca/realm/}.
\end{abstract}

\section{Introduction}
\label{sec:intro}
\begin{figure}[t!]
\centering
  \centering
  \includegraphics[width=\linewidth]{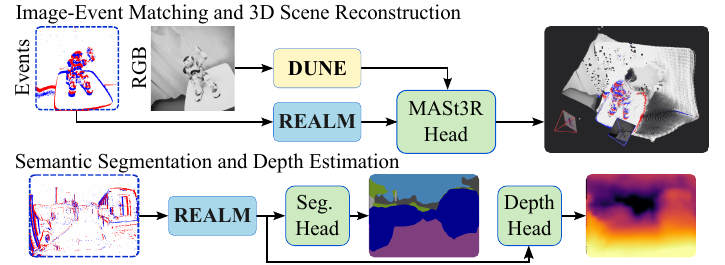}
\caption{\textbf{Overview of \methodname's cross-modal versatility.} 
Our model enables diverse downstream applications, which include feature matching using the frozen MASt3R~\cite{Mast3r2025eccv} decoder and dense prediction tasks like semantic segmentation and depth estimation using simple linear heads.}
\label{fig:eyecatcher}
\vspace*{-5mm}
\end{figure}

Visual perception is fundamental to modern computer vision and intelligent systems, driving progress in applications ranging from autonomous navigation and augmented reality to computational photography. The vast majority of these systems rely on conventional RGB cameras, which capture dense frames at fixed intervals. While effective in controlled environments, frame-based sensors suffer from fundamental limitations: motion blur, limited dynamic range, and high latency~\cite{gallego2022tpami}. These constraints degrade performance in challenging conditions, such as high-speed motion~\cite{shariff2024event} or extreme lighting environments~\cite{kong2024openess}.

Event cameras are bio-inspired sensors that have found several applications in robotics and computer vision~\cite{polizzi2025vibes, botao2024microsaccade} and offer a paradigm shift in visual sensing. By asynchronously recording per-pixel brightness changes, event cameras produce sparse event streams with microsecond-scale latency and high dynamic range~\cite{gehrig2024low}. This sensing principle reduces motion blur and drastically lowers power consumption, making event data distinctively advantageous for analyzing dynamic scenes~\cite{messikommer2023data, Rosinol_2018_RAL}.

Despite these advantages, learning task-agnostic feature representations from event data remains an open challenge.
Event streams are sparse, asynchronous, and structurally distinct from images, posing significant challenges for the direct application of standard computer vision techniques. 
Furthermore, unlike the RGB domain---which benefits from large-scale datasets---labelled event datasets are scarce and small in scale. Consequently, existing event-based models are typically trained from scratch for narrow, specific tasks (e.g., for optical flow computation~\cite{Shiba22eccv} or classification~\cite{deng2022cvpr} only), failing to generalize across different downstream applications.

In the RGB domain, the landscape of computer vision techniques has been reshaped by large-scale vision transformers (ViTs). 
Foundation models such as the DINO family~\cite{caron2021dino, oquab2024dinov2, simeoni2025dinov3} or DUNE~\cite{Sariyildiz2025CVPR} have demonstrated that pretraining on massive image collections yields universal encoders that transfer seamlessly to diverse tasks. 
However, the event vision community has largely been excluded from this revolution due to the significant modality gap between intensity-based frames and temporally-driven event streams.

To bridge this gap, we propose \methodname, a cross-modal visual encoder that unifies RGB and event representations within a shared latent space (see~\autoref{fig:eyecatcher}). 
Building upon the DUNE~\cite{Sariyildiz2025CVPR} architecture, \methodname introduces a lightweight, modality-specific input embedder and leverages Low-Rank Adaptation (LoRA)\cite{hu2022lora} to align event features with the semantically rich and geometrically consistent RGB manifold. Crucially, this approach allows us to align the modalities without retraining the backbone, preserving the rich semantic and geometric knowledge learned from millions of RGB images. This design enables both intra-modal (event–event) and cross-modal (RGB–event) feature matching, facilitating modality-invariant transfer learning where features remain spatially and semantically consistent regardless of the input sensor.
Unlike cross-modal distillation between dense and synchronous sensor modalities such as depth, LiDAR, or thermal, the event modality poses challenges intrinsic to how events are generated. 
For example, regions of static brightness produce no events, leaving parts of the scene unobserved. 
The model must therefore learn dense structure without hallucinating content where no signal exists. 
At the same time, a uniform distillation objective should be avoided as it would penalize the student over regions it cannot observe.
Addressing these two properties is what allows a frozen, image-trained decoder to operate zero-shot on event features.

We validate \methodname across multiple datasets on three distinct downstream tasks: monocular depth estimation, semantic segmentation, and wide-baseline feature matching. For the latter, we specifically evaluate cross-modal (RGB–event) and intra-modal (event–event) correspondences, demonstrating that our aligned manifold allows a frozen, image-trained matching head to function across disparate sensors without retraining.
Our evaluation focuses on two core hypotheses: first, that a shared latent space allows event-based data to leverage knowledge from large-scale RGB foundation models. Second, that this alignment is sufficiently precise to enable the direct, zero-shot\footnote{Throughout this work, \textit{zero-shot} refers to a head trained exclusively on features from a frozen DUNE~\cite{Sariyildiz2025CVPR} encoder and deployed directly onto event features from REALM, with no event-specific supervision, fine-tuning, or adaptation at any stage. }
application of frozen image-trained decoders to asynchronous event streams.
While we demonstrate that the aligned latent space is rich enough to support dense prediction tasks like depth and segmentation using only simple linear heads, our most significant results lie in wide-baseline feature matching. 
Here, \methodname enables the direct use of frozen, RGB-trained geometric heads, such as MASt3R, to perform cross-modal and intra-modal matching. In this challenging domain, our method not only achieves state-of-the-art performance but does so by outperforming specialized event-only baselines.

\noindent We make the following contributions.
\begin{itemize}
    \item We introduce \methodname, a unified encoder that aligns event and RGB modalities in a shared, semantically rich latent space.
    \item We demonstrate that our alignment is sufficiently precise to enable the direct, zero-shot application of frozen, image-trained geometric decoders to event streams, establishing a new state-of-the-art in event-based feature matching and pose estimation.
    \item We show that lightweight LoRA~\cite{hu2022lora} fine-tuning bridges the modality gap without catastrophic forgetting. The resulting frozen encoder generalizes to diverse tasks, from dense depth estimation to semantic segmentation, providing a data-efficient solution to the scarcity of annotated event datasets.
    \item We provide an open-source implementation of \methodname, together with models, at \href{https://papers.starslab.ca/realm}{https://papers.starslab.ca/realm/}.
\end{itemize}

\section{Related Work}
\label{sec:related}
We first cover event representations, then review the three core tasks our method addresses: semantic segmentation, depth estimation, and feature matching. Finally, we provide an overview of cross-modal knowledge transfer.

\subsection{Event Representation}
Event cameras asynchronously record changes in scene brightness, producing a sparse stream of events that depends on relative motion between the camera and the scene~\cite{polizzi2025vibes,botao2024microsaccade}.
Unlike conventional frames, event data encode temporal contrast rather than absolute intensity, which poses challenges for standard deep networks that assume dense, grid-structured inputs.

To bridge the modality gap, several representations have been proposed to convert event streams into tensor formats suitable for deep networks~\cite{Gehrig2019ICCV}. 
Common representations include voxel grids~\cite{zhu2019cvpr}, which bin events temporally, and Tencode~\cite{huang2023wacv}, which renders a chunk of events as an RGB image, using the red/blue channels for event polarity and the green channel for the relative timestamp. More recently, ERGO~\cite{Zubic2023ICCV} optimizes the event ordering for detection tasks.

\subsection{Semantic Segmentation}
Semantic segmentation assigns a class label to each pixel in the camera frame, providing dense scene understanding essential for robotics and autonomous driving.
In the RGB domain, while early methods relied on CNN-based encoder-decoders like SegNet~\cite{badrinarayanan2017segnet}, the field has largely shifted toward Transformer-based architectures that leverage large-scale pretraining for robust pixel-level classification~\cite{cheng2022masked, xie2021segformer, kirillov2023segment}. An in-depth review of visual segmentation methods is in~\cite{thisanke2023semantic}.

For event cameras, segmentation is complicated by the sparsity and lack of texture in the data. 
Early works~\cite{alonso2019cvpr} addressed this by adapting CNN architectures and releasing the first event-segmentation benchmarks. Similarly, ESS~\cite{sun2022ess} proposed a recurrent encoder-decoder to maintain temporal consistency in predictions.
More recently, ESEG~\cite{ESEG2025AAAI} tackles the sparsity problem by introducing explicit edge-semantic supervision to locate reliable cues.
Alternative approaches leverage RGB data to guide learning: EvDistill~\cite{Wang2021CVPR} uses a teacher–student framework to transfer knowledge from frames to events, while HALSIE~\cite{das2024halsie} fuses both modalities to extract rich spatiotemporal features.

However, these methods often require training complex decoders or distillation pipelines from scratch. 
In contrast, \methodname demonstrates that by aligning event representations with the frozen DUNE~\cite{Sariyildiz2025CVPR} latent space, we achieve dense semantic maps using only a simple linear head, inheriting strong scene understanding capabilities without the need for extensive task-specific supervision.

\subsection{Depth Estimation}
Depth estimation predicts per-pixel distances, a critical capability for navigation, mapping, and 3D reconstruction. 
For RGB images, Eigen et al.~\cite{eigen2014depth} originally formulated depth estimation as a continuous regression task using CNNs. However, recent state-of-the-art approaches have recast depth estimation as a classification problem (predicting depth bins), leveraging the global context of Transformers~\cite{bhat2021cvpr, yuan2022cvpr}. Foundation models like DINOv2~\cite{oquab2024dinov2}, MASt3R~\cite{Mast3r2025eccv}, VGGT~\cite{wang2025vggt}, and DepthAnything~\cite{Yang2024depthanything} have further demonstrated that general-purpose features can yield high-quality depth maps without specialized depth architectures.

For event data, monocular depth estimation is complicated by sparsity and motion dependence, as static regions generate no events and create "blind spots" that necessitate temporal aggregation. Previous works have addressed these challenges by employing recurrent U-Nets to aggregate temporal information or by fusing sparse events with frames to densify the resulting predictions~\cite{hidalgo20203dv, gehrig2021RAL}. Furthermore, self-supervised methods have emerged to bypass the need for ground-truth depth labels by instead leveraging cross-modal consistency between synchronized event streams and intensity frames~\cite{zhu2023iros, zhu2019cvpr}. A more comprehensive review on depth estimation with event cameras is given by~\cite{ghosh2025tpami}.

Unlike these approaches, which often rely on complex recurrent architectures to "fill in" missing data, we adopt a classification-based paradigm and demonstrate that aligning an event encoder with the geometrically rich DUNE~\cite{Sariyildiz2025CVPR} latent space is sufficient to produce dense depth maps, effectively bypassing the need for specialized event-depth architectures.

\subsection{Feature Matching and Localization}
Feature matching under wide-baseline and cross-modal conditions is a cornerstone for tasks such as visual odometry, SLAM, and 3D reconstruction.

The literature on feature extraction and matching for visual cameras is extensive, ranging from classical hand-crafted descriptors~\cite{lowe2004sift, rublee2011orb, bay2008surf, leutenegger2011brisk} to modern learning-based methods~\cite{detone2018superpoint, Sarlin20cvpr, Yi16eccv}. 
More recent approaches leverage geometric priors to improve feature stability and robustness across viewpoints~\cite{wang2025vggt, Mast3r2025eccv, dust3rcvpr24}.

In contrast, event-based feature matching remains challenging because the absence of absolute intensity cues makes constructing repeatable descriptors difficult.
Existing methods typically rely on spatio-temporal correlations or event accumulation strategies to detect and match features~\cite{messikommer2023data, rebecq2017bmvc}. 
Recent work~\cite{huang2023wacv} introduced a novel event representation, Tencode, which learns feature correspondences by leveraging matches from a pre-trained network such as SuperPoint~\cite{detone2018superpoint}. Other methods, such as MINIMA~\cite{ren2025minima}, develop modality-invariant matching by utilizing a generative data engine to synthesize massive multimodal datasets from RGB pairs, enabling cross-modal generalization through fine-tuning.

Our approach differs from these paradigms by leveraging the latent space of a foundation model to incorporate 3D geometric priors. \methodname enables the zero-shot use of frozen image-trained heads~\cite{Mast3r2025eccv} for matching across both modalities and viewpoint changes. This formulation provides a unified geometric embedding space that facilitates seamless cross-modal correspondence.

\subsection{Cross-Modal Knowledge Transfer}
Recent successes in event-to-frame reconstruction, such as E2VID~\cite{Rebecq19pami}, have demonstrated that event streams contain sufficient information to recover high-fidelity photometric representations of a scene.
While methods like EvDistill~\cite{Wang2021CVPR} leverage this by mapping events into the image domain before utilizing large-scale visual models, this intermediate reconstruction step often introduces computational overhead and potential artifacts. 
We therefore posit that it is possible to bypass image reconstruction entirely by directly mapping events into the latent representation space of large, pretrained visual encoders.

\section{Methodology}
\label{sec:methodology}
In this section, we present our approach for training and refining \methodname. We first provide background on the DUNE\cite{Sariyildiz2025CVPR} encoder, followed by a formal problem definition for cross-modal alignment. Finally, we detail our architectural modification strategy, focusing on the adaptation of a frozen foundation model for asynchronous event data.

\subsection{Background: DUNE}
\label{sec:dune_desc}

Our work builds upon the DUNE~\cite{Sariyildiz2025CVPR} framework, a universal vision transformer (ViT) designed to unify diverse 2D and 3D perception tasks into a single encoder through multi-teacher distillation.

DUNE~\cite{Sariyildiz2025CVPR} is trained using a set of $N$ heterogeneous teacher models $\mathcal{T} = \{\mathcal{T}_1, \dots, \mathcal{T}_N\}$. Each teacher $\mathcal{T}_i$ is parameterized by its own encoder $t_i(x)$, which maps an input image $x$ into a set of feature vectors $Z \in \mathbb{R}^{(HW+1) \times d}$. These features include $HW$ patch tokens and an optional global CLS token.
The goal of distillation is to learn a student encoder $f(x)$ such that its representations, after a teacher-specific projection, match those of the teachers. For each teacher $\mathcal{T}_i$, DUNE~\cite{Sariyildiz2025CVPR} employs a teacher-specific transformer projector $h_i$ to map the student’s features into the teacher's latent space.

DUNE~\cite{Sariyildiz2025CVPR} distills from three distinct teachers: DINOv2~\cite{oquab2024dinov2}, MASt3R~\cite{Mast3r2025eccv}, and Multi-HMR~\cite{baradel2024mhr}. After the student encoder $f$ is trained, the task-specific decoders (or heads) are fine-tuned independently while keeping the encoder frozen:
$y_{task} = \text{Decoder}_{task}(f(x))$.
This decoupled training ensures that the student encoder retains a ``universal'' latent manifold suitable for diverse downstream applications without needing to store or compute teacher-specific projectors during inference.

In \methodname, we leverage the open-source MASt3R~\cite{Mast3r2025eccv} head, which was refined on DUNE's universal features. We utilize this head directly in \methodname without any event-based refinement, providing a rigorous zero-shot benchmark for our cross-modal alignment strategy.

We emphasize that our alignment framework is not tied to DUNE~\cite{Sariyildiz2025CVPR} specifically: any ViT-based encoder with a sufficiently general latent space could serve as the teacher. We adopt DUNE~\cite{Sariyildiz2025CVPR} because its multi-teacher distillation yields a manifold that already spans semantic, geometric, and human-centric priors, which is well matched to the diverse downstream tasks we target.

\subsection{Problem Definition}
\label{sec:prob_def}
\begin{figure*}[t!]
\centering
\includegraphics[width=\linewidth]{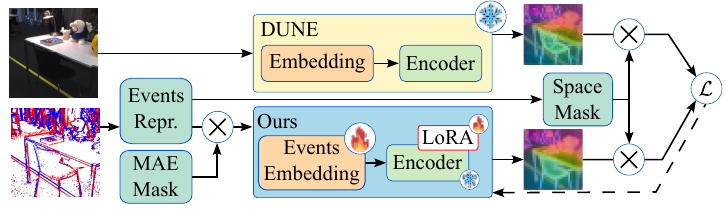}
\vspace*{-4mm}
\caption{\textbf{Overview of the cross-modal distillation framework.} Event representations undergo masked autoencoders (MAE)-style dropout before being processed by a trainable embedding layer and a LoRA-adapted student encoder. %
A progressive spatial mask is applied to the distillation loss ($\mathcal{L}$) to focus the alignment on regions with active event data, preventing background overfitting.}
\label{fig:training}
\vspace*{-1mm}
\end{figure*}

Our primary objective is to learn a mapping $f : \mathcal{E} \rightarrow \mathcal{Z}_E$ that projects an event representation $e \in \mathcal{E}$ into a latent feature vector $\mathbf{Z}_E \in \mathcal{Z}_E$. We optimize $f$ to minimize the distance between $\mathcal{Z}_E$ and $\mathcal{Z}_I$, where $\mathcal{Z}_I$ represents the latent manifold of the DUNE~\cite{Sariyildiz2025CVPR} foundation model. As described in~\autoref{sec:dune_desc}, for a given image $x$, the corresponding reference feature is defined as $\mathbf{Z}_I = f_{\text{DUNE}}(x)$, where $\mathbf{Z}_I \in \mathbb{R}^{(HW+1) \times d}$.

To select the optimal input format for $e$, we evaluated various representations (see~\autoref{sec:related}) and found that the choice of input has a negligible effect on final alignment quality (see ~\autoref{sec:results}). Consequently, we adopt the voxel grid due to its wide adoption in event-based learning, compatibility with established baselines, and its suitability for the convolutional embedder design we employ, see Appendix~\autoref{app:models}.

\subsection{Training and Architecture}
\label{sec:method}

Standard vision transformer (ViT) architectures, including DUNE, are designed to process RGB images through a patchified embedding stage.
Formally, the DUNE~\cite{Sariyildiz2025CVPR} encoder can be decomposed as $f_{\text{DUNE}} = f_{\text{back}} \circ f_{\text{emb}}$, where $f_{\text{emb}}: \mathbb{R}^{H \times W \times C} \rightarrow \mathbb{R}^{M \times d}$ projects raw pixels into $M$ patch tokens, and $f_{\text{back}}: \mathbb{R}^{M \times d} \rightarrow \mathcal{Z}_I$ represents the transformer backbone.
\paragraph{Event-Specific Embedding.} Because our voxel grid representation contains five temporal bins and distinct structural properties compared to RGB frames, the original $f_{\text{emb}}$ is incompatible. 
We substitute the image-based patch module with $f_{\text{emb}}^{E}$, a lightweight convolutional stem that maps the voxel grid directly into a set of $M \times d$ tokens. 
Our goal is not to force the intermediate embeddings to match ($f_{\text{emb}}(x) = f_{\text{emb}}^{E}(e)$), but rather to ensure that the final latent outputs are aligned within the foundation manifold: $f_{\text{back}}(f_{\text{emb}}(x)) \approx f_{\text{back}}^{E}(f_{\text{emb}}^{E}(e))$.
\paragraph{Low-Rank Adaptation (LoRA).} To adapt the backbone $f_{\text{back}}$ for the event modality while preserving its pretrained semantic and geometric knowledge, we employ Low-Rank Adaptation (LoRA)~\cite{hu2022lora}. We keep the original weights $W \in \mathbb{R}^{d \times k}$ frozen and model the necessary cross-modal refinement as a low-rank decomposition: $W' = W + AB$, where $A \in \mathbb{R}^{d \times r}$ and $B \in \mathbb{R}^{r \times k}$ are trainable matrices with rank $r \ll \min(d, k)$. This allows \methodname to bridge the modality gap without the risk of catastrophic forgetting.

To robustly distill knowledge from the DUNE~\cite{Sariyildiz2025CVPR} encoder, we propose a dual-masking training strategy described as follows (see~\autoref{fig:training}).

\begin{figure}[t!]
\centering
  \centering
  \includegraphics[width=\linewidth]{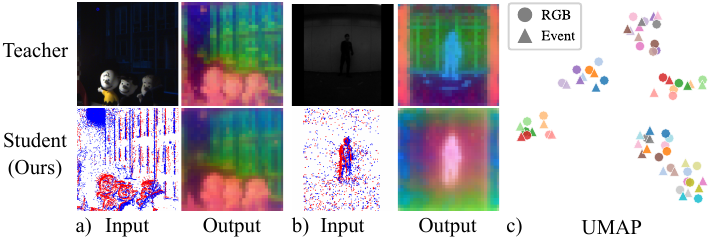}
\caption{\textbf{Qualitative comparison of DUNE and \methodname feature manifolds.} The spatial feature maps in a) and b), visualized via PCA, show that \methodname %
learns to %
identify semantic structures similarly to the RGB network. In b), 
the masking suppresses static background visible to the camera but invisible to the events.
In c), the UMAP plot further confirms this by showing event and RGB features form consistent clusters.
}
\label{fig:features}
\vspace*{-5mm}
\end{figure}

\paragraph{Progressive Spatial Distillation.} 
To prevent the student model from overfitting to static backgrounds (see~\autoref{fig:features}b), we constrain the distillation loss using a spatially-aware mask $\mathcal{M}$ derived from the input event activity.

Let $E^i \in \{0,1\}^{M}$ be a binary occupancy grid for the voxel grid input $e_i$, where $E^i_j=1$ if the $j$-th voxel contains at least one event, and $E^i_j=0$ otherwise. We define the active distillation mask for $e_i$ at training step $s$ as: $\mathcal{M}^i(s) = \text{Dilation}(E^i, \sigma(s))$, where $\text{Dilation}(\cdot, \sigma)$ denotes a morphological dilation operation with a structuring element of radius $\sigma$. We introduce a spatial curriculum by defining $\sigma(s)$ as a monotonically increasing function of the training progress: $\sigma(s) = \min\left(\sigma_{max}, \lfloor \alpha \cdot s \rfloor \right)$, where $\alpha$ is the expansion rate and $\sigma_{max}$ is the radius required to cover the entire spatial grid.

By gradually expanding the penalization region $\mathcal{M}$, we force the model to move beyond trivial edge-matching and learn a holistic scene understanding by leveraging the teacher's static priors in regions initially devoid of events.

\paragraph{Input Patch Masking.} 
To encourage the network to build compressed global representations and discourage reliance on local interpolation, we employ a stochastic patch dropout strategy inspired by Masked Autoencoders (MAE)~\cite{he2022mae}.
Following an initial warmup phase, we uniformly mask a fixed percentage $\rho$ of the input event tokens $\mathcal{X}_E=f_{\text{emb}}^E(e)$ produced by the embedder. By denying the backbone access to the full spatial grid, we force the LoRA-adapted layers to reconstruct dense, cross-modal features from highly degraded inputs. This prevents the model from relying on local interpolation and ensures the learned latent space $Z_E$ captures the geometric and semantic priors of the DUNE teacher.

\paragraph{Loss Function.} To ensure feature alignment, we define the total loss as $\mathcal{L}_{\text{total}}=\sum_{i=1}^B \mathcal{L}_{i}$, where $\mathcal{L}_{i}$ is the loss for the voxel grid $i$ in a batch of size $B$ defined as a weighted combination of $\ell_1$ distance, cosine similarity and MSE losses between the DUNE~\cite{Sariyildiz2025CVPR} embeddings of the image and event modalities:
\begin{equation}
\mathcal{L}^i = \frac{1}{\sum_{j=1}^{M} \mathcal{M}^i_j(s)} \sum_{j=1}^{M} \mathcal{M}^i_j(s) \cdot \left[ \lambda_{\text{MSE}} \mathcal{L}_{\text{MSE}}^j + \lambda_{\text{cos}} \mathcal{L}_{\text{cos}}^j + \lambda_{\ell_1} \mathcal{L}_{\ell_1}^j \right]
\end{equation}
where $\lambda_{\text{MSE}}$, $\lambda_{\text{cos}}$, and $\lambda_{\text{L1}}$ control the relative contributions of each term.
This objective enforces per-dimension and global directional alignment, promoting cross-modal generalization as shown by the uniform manifold approximation and projection (UMAP)~\cite{McInnes2018umap} in~\autoref{fig:features} c).
We show the effect of the proposed masking strategy in Appendix \autoref{app:training}.

\section{Results}
\label{sec:results}
In this section, we present a comprehensive experimental evaluation of REALM across three primary downstream tasks: monocular depth estimation, semantic segmentation, and wide-baseline feature matching. We report how our model generalizes to other tasks such as image reconstruction in~\autoref{app:extra_tasks}.

To train the event-specific embedder and the LoRA adapters, we utilize a diverse collection of synchronized event-RGB datasets. 
We selected DSEC~\cite{gehrig2021dsec, Gehrig3dv2021, kong2024openess},  EventScape~\cite{gehrig2021RAL}, EventPointMesh~\cite{Hori2025TVCG}, EDS~\cite{hidalgo2022cvpr}, and M3ED~\cite{chaney2023cvpr} to provide a comprehensive overview of diverse geometric and semantic structures, covering a wide range of scenarios.
All training input data are resized and cropped to a voxel-grid with a resolution of 448$\times$448 with five temporal bins.
For further details on the datasets and our data preparation, refer to the Appendix~\autoref{app:datasets}.

To choose the event representation for our model, we conducted a preliminary evaluation on 5\% of our training data comparing voxel grids~\cite{zhu2019cvpr}, Tencode~\cite{huang2023wacv}, and ERGO~\cite{Zubic2023ICCV}. All three yielded nearly identical alignment losses (0.0687, 0.0698, and 0.0698, respectively), confirming that the choice of input representation has a negligible effect on cross-modal alignment quality. 

We apply LoRA to the attention, projection, and feed-forward layers with a rank of 32 and a learning rate of $1\times10^{-3}$ following a cosine decay schedule.

For each task, we compare \methodname against state-of-the-art architectures specifically engineered for event data. We also report the performance of DUNE \cite{Sariyildiz2025CVPR} on synchronized RGB frames to compare \methodname to its teacher and show where the event modality helps in perception tasks.

A key strength of our framework is the minimal overhead required for downstream task adaptation:
with the segmentation and monocular depth tasks, we demonstrate that \methodname's latent space is semantically and geometrically rich enough to support these tasks using only simple linear heads. Notably, these heads are not trained on events, but rather on RGB images on the frozen DUNE~\cite{Sariyildiz2025CVPR} encoder. See the Appendix~\autoref{app:models} for further details on the models.

In the same way, for feature matching, we directly utilize the frozen MASt3R\cite{Mast3r2025eccv} head that has been trained on a large dataset of 1.7M images~\cite{Sariyildiz2025CVPR}. This requires no retraining or fine-tuning on event data, serving as a strong verification of our cross-modal alignment strategy.

\subsection{Depth Estimation}
\label{sec:depth_estimation}
\begin{table*}[t!]
\centering
\setlength{\tabcolsep}{2.9pt}
    \begin{tabular}{c|c|c|cccc}
        \toprule
        \multicolumn{2}{c|}{} & \multicolumn{5}{c}{Avg. Absolute Depth Err. [m]} \\
        \midrule
        \multirow{2}{*}{\textbf{Dataset}} & 
        \textbf{Depth} &
        \textbf{DUNE} & 
        \textbf{E2Depth} & 
        \textbf{Zhu et al.} & 
        \textbf{EMoDepth} & 
        \textbf{\methodname} \\
        & \textbf{Cut-offs} & \cite{Sariyildiz2025CVPR} & \cite{hidalgo20203dv} & \cite{zhu2019cvpr} & \cite{zhu2023iros} & (Ours) \\
        \toprule
        \multirow{3}{*}{outdoor day 1} 
        & 10 m & 1.16 & \ul{1.85} & 2.72 & \textbf{1.40} & \ul{1.85} \\
        & \colortab 20 m & \colortab 1.76 & \colortab 2.64 & \colortab 3.84 & \colortab \textbf{2.07} & \colortab \ul{2.42} \\
        & 30 m & 2.12 & 3.13 & 4.40 & \textbf{2.65} & \ul{2.76} \\
        \hline
        \multirow{3}{*}{outdoor night 1} 
        & 10 m & 2.13 & 3.38 & 3.13 & \ul{2.18} & \textbf{2.08} \\
        & \colortab 20 m & \colortab 3.10 & \colortab 3.82 & \colortab 4.02 & \colortab \ul{2.70} & \colortab \textbf{2.51} \\
        & 30 m & 3.51 & 4.46 & 4.89 & \ul{3.64} & \textbf{3.18} \\
        \hline
        \multirow{3}{*}{outdoor night 2} 
        & 10 m & 2.45 & \textbf{1.67} & 2.19 & 2.06 & \ul{2.00} \\
        & \colortab 20 m & \colortab 3.46 & \colortab \ul{2.63} & \colortab 3.15 & \colortab 2.76 & \colortab \textbf{2.31} \\
        & 30 m & 3.86 & 3.58 & 3.92 & \ul{3.42} & \textbf{2.98} \\
        \hline
        \multirow{3}{*}{outdoor night 3} 
        & 10 m & 2.33 & \textbf{1.42} & 2.86 & 2.09 & \ul{1.79} \\
        & \colortab 20 m & \colortab 3.37 & \colortab \ul{2.33} & \colortab 4.46 & \colortab 2.82 & \colortab \textbf{2.15} \\
        & 30 m & 3.79 & \ul{3.18} & 5.05 & 3.52 & \textbf{2.97} \\
        \bottomrule
    \end{tabular}
\vspace{2mm}
\caption{\textbf{Quantitative results on the MVSEC~\cite{zhu2018mvsec} dataset.} 
We report results for DUNE~\cite{Sariyildiz2025CVPR} as a reference, and for our event-only method, \methodname. Comparisons are provided against a supervised approach~\cite{hidalgo20203dv}, as well as unsupervised methods~\cite{zhu2019cvpr,zhu2023iros}.}
\label{tab:depth_mvsec_results}
\vspace*{-6mm}
\end{table*}

\begin{table*}[t!]
\centering
\resizebox{\textwidth}{!}{%
\renewcommand{\arraystretch}{1.05}
\begin{tabular}{l|ccccc|ccc|ccc|cc}
\textbf{Method} & \textbf{AbsR} $\downarrow$ & \textbf{SqR} $\downarrow$ & \textbf{RMSE} $\downarrow$ & \textbf{log} $\downarrow$ & \textbf{SILog} $\downarrow$ & $\mathbf{\sigma_1}$ $\uparrow$ & $\mathbf{\sigma_2}$ $\uparrow$ & $\mathbf{\sigma_3}$ $\uparrow$ & 10 & 20 & 30 & \textbf{Latency} (ms) $\downarrow$ & \textbf{Params} $\downarrow$  \\ \hline
DUNE  & 0.351 & 0.424 & 7.828 & 0.415 & 0.166 & 0.559 & 0.764 & 0.878 & 2.02 & 2.92 & 3.32
& 6.02 (A100) & 0.39M \\ \hline
\colorrow EReFormer\cite{liu2024ereformer}& \tb{0.275} & \ul{0.382} & -- & -- & \tb{0.120} & \tb{0.607} & \tb{0.807} & \tb{0.915} & \tb{1.38} & \ul{2.15} & \tb{2.73}
& 35.17 (Tesla V100) & 29.9M \\ 
D. AnyEv.~\cite{bartolomei2025iccv}& 0.362 & 0.697 & \tb{6.511} & \ul{0.438} & 0.211 & 0.494 & \ul{0.760} & \ul{0.890} & -- & -- & --
& \ul{9.20} (A100) & \ul{4.06M} \\ 
\colorrow AnyEv.\ Stream~\cite{zhu2025iccv}& -- & -- & -- & -- & -- & -- & -- & --  & \tb{1.38} & \tb{2.02} & 3.01
& 22.00 (A100) & 10.89M \\ 
REALM (Ours)  & \ul{0.349} & \tb{0.312} & \ul{8.109} & \tb{0.421} & \ul{0.169} & \ul{0.500} & 0.754 & 0.875 & 1.93 & 2.35 & \ul{2.97}
& \tb{6.71 (A100)} & \tb{0.39M} \\ 
\end{tabular}%
  }
\vspace*{1mm}
\captionof{table}{\tb{Extended Depth Metrics and Comparison on MVSEC.} Specialized transformer decoders reach strong accuracy at substantially higher parameter and latency cost, whereas REALM remains competitive with only a linear head on a frozen backbone. Params are decoder/head parameters only. Values are averaged over scenes.}
\label{tab:depth_mvsec_results_extended}
\vspace*{-4mm}
\end{table*}

\begin{figure}[t!]
\centering
  \centering
  \includegraphics[width=\linewidth]{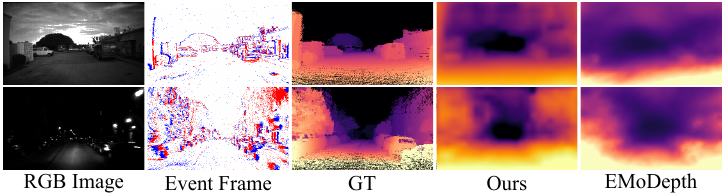}
\caption{\textbf{Qualitative depth estimation results on the MVSEC~\cite{zhu2018mvsec} dataset.} From left to right: RGB frame (only for visualization purposes), event frame, ground truth depth, and \methodname and EMoDepth~\cite{zhu2023iros} predictions. Despite the sparsity of the event stream, \methodname preserves the scene structure just using a single linear projector.}
\label{fig:depth_estimation}
\vspace*{-5mm}
\end{figure}

We evaluate the dense geometric reasoning capabilities of \methodname by performing monocular depth estimation. 
Following recent trends in foundation model evaluation~\cite{oquab2024dinov2}, we aim to demonstrate that a frozen \methodname encoder, paired with a minimal prediction head, can compete with specialized event-based depth architectures.
Accordingly, our setup is minimal, as we utilize a single linear head trained exclusively on frozen RGB features from DUNE~\cite{Sariyildiz2025CVPR}, with no event-specific depth supervision at any stage.
\paragraph{Experimental Setup.} We benchmark our method on the MVSEC~\cite{zhu2018mvsec} dataset using the evaluation protocol established by E2Depth~\cite{hidalgo20203dv}. We treat depth estimation as a classification task by discretizing depth values into bins and predicting per-pixel probability distributions. 
To verify the robustness of our latent space, we first train a linear head for 300 epochs on top of the frozen DUNE~\cite{Sariyildiz2025CVPR} backbone using real and synthetic RGB images, from MVSEC~\cite{zhu2018mvsec} and DENSE~\cite{hidalgo20203dv}, respectively. 
Then we use the trained head directly on \methodname event features with no finetuning. 
Further details on the training are in the Appendix~\autoref{app:heads_training}. 

To accommodate the $260 \times 346$ resolution of the MVSEC~\cite{zhu2018mvsec} dataset while maintaining the $448 \times 448$ input size of the foundation model, we symmetrically pad the original data. We then remove the padding from the prediction. To ensure temporal stability, we implement a memory hold mechanism. That is, in time windows where no event activity is recorded, the model holds the most recent valid depth estimate, preventing noisy outputs during brief sensor inactivity.

\paragraph{Quantitative Results.} The results are summarized in~\autoref{tab:depth_mvsec_results} and~\autoref{fig:depth_estimation}.
Despite this significant gap in both architectural complexity and task-specific supervision, \methodname remains competitive across nearly all evaluated sequences.
Notably, our approach demonstrates superior robustness in challenging outdoor night scenarios where traditional frame-based sensors degrade. For instance, in the outdoor night 1 sequence at a 20\,m cut-off, \methodname achieves an error of 2.51\,m, significantly outperforming both the unsupervised EMoDepth~\cite{zhu2023iros} (2.70\,m) and Zhu et al.~\cite{zhu2019cvpr} (4.02\,m).
Crucially, although \methodname is trained only on RGB data, distilled from the frame-based teacher DUNE~\cite{Sariyildiz2025CVPR}, it surpasses the teacher model in low-illumination and high-dynamic-range conditions, that is, in the regime where events are more informative than frames, which lose detail in under- and over-exposed regions. The student can thus exceed the source of its own supervision: our aligned manifold not only inherits DUNE's representation but also lets the event modality contribute cues unavailable in RGB.

In~\autoref{tab:depth_mvsec_results_extended} we report additional depth metrics alongside recent transformer-based methods~\cite{liu2024ereformer, bartolomei2025iccv, zhu2025iccv}. While these specialized decoders achieve strong accuracy on depth estimation, they do so with $10$--$77\times$ more parameters and $1.4$--$5\times$ higher latency than \methodname, which attains competitive results using only a linear head on a frozen backbone.

\subsection{Semantic Segmentation}
\label{sec:semantic_seg}
\begin{table}[t]
\centering
\setlength{\tabcolsep}{4pt} %
    \begin{tabular}{l|c|ccccc}
        \toprule
        \textbf{Metric} & \textbf{DUNE} & \textbf{ESS} & \textbf{EV-SegNet} & \textbf{HALSIE} & \textbf{ESEG} & \textbf{\methodname} \\
         & \cite{Sariyildiz2025CVPR} & \cite{sun2022ess} & \cite{alonso2019cvpr} & \cite{das2024halsie} & \cite{ESEG2025AAAI} & (Ours) \\
        \midrule
        \textbf{Acc. [\%]} $\uparrow$ & 93.33 & \underline{89.25} & 88.61 & 89.01 & \textbf{91.47} & 89.23 \\
        \colorrow \textbf{mIoU [\%]} $\uparrow$ & 67.64 & 51.57 & 51.76 & 52.43 & \textbf{57.55} & \underline{55.37} \\
        \bottomrule
    \end{tabular}
\vspace{2mm}
\caption{\textbf{Quantitative results on the DSEC dataset.} 
We use a linear segmentation head to perform the segmentation task. DUNE shows the upper-bound performance of the shared architecture on intensity frames.}
\label{tab:segmentation_evaluation}
\vspace*{-4mm}
\end{table}

\begin{figure}[t!]
\centering
  \centering
  \includegraphics[width=\linewidth]{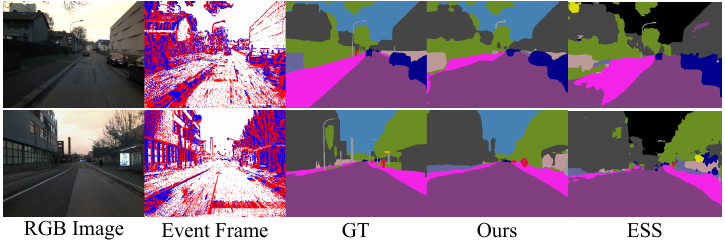}
\caption{\textbf{Qualitative semantic segmentation on the DSEC~\cite{gehrig2021dsec} driving dataset.} \methodname effectively identifies key classes such as road, vehicles, and sidewalk. Our model achieves these results using a frozen backbone and a single linear head, demonstrating the semantic richness of the aligned latent space.}
\label{fig:sem_seg}
\vspace*{-5mm}
\end{figure}

\begin{figure}[t!]
\centering
  \centering
  \includegraphics[width=\linewidth]{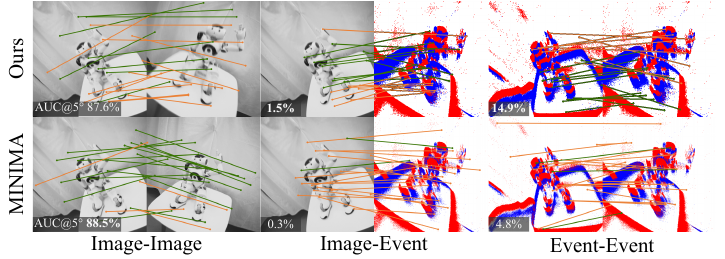}
\caption{\textbf{Cross-modal and intra-modal feature matching.} We report the AUC at 5$^\circ$ per scene and visualize the correspondence estimation results for inliers and outliers in \green{green} and \orange{orange}, respectively. 
}
\label{fig:matches_visual}
\vspace*{-5mm}
\end{figure}

To evaluate the semantic qualities of the aligned latent space, we benchmark \methodname on the task of semantic segmentation. This experiment serves to demonstrate that the high-level scene understanding present in RGB foundation models can be seamlessly transferred to the event modality.

\paragraph{Experimental Setup.} We utilize the DSEC~\cite{gehrig2021dsec} dataset, which provides semantic labels across eleven distinct classes. Following the evaluation protocol of ESS~\cite{sun2022ess}, we compute standard confusion-matrix-based metrics, including Pixel Accuracy and Mean Intersection over Union (mIoU). To highlight the off-the-shelf utility of our features, we employ a minimal single-layer linear head. The head is trained using the frozen DUNE~\cite{Sariyildiz2025CVPR} encoder on RGB images and subsequently used as is, with no further finetuning, on \methodname event embeddings.

To accommodate the $480 \times 640$ resolution of the DSEC~\cite{gehrig2021dsec} dataset while maintaining the $448 \times 448$ input size of the foundation model, we implement an overlapping tiling strategy. During inference, we extract four $448 \times 448$ tiles from the corners of the input grid. The final per-pixel prediction is determined by accumulating the raw model outputs across tiles and normalizing the result by a count mask to average overlapping regions. To ensure temporal stability, we implement a memory hold mechanism as described in~\autoref{sec:depth_estimation}.
\smallskip

\paragraph{Quantitative Results.} The results are summarized in~\autoref{tab:segmentation_evaluation} and~\autoref{fig:sem_seg}. 
\methodname achieves a Pixel Accuracy of 89.23\% and an mIoU of 55.37\%. 
These results are highly competitive with specialized architectures such as ESS~\cite{sun2022ess} (51.57\% mIoU) and EV-SegNet~\cite{alonso2019cvpr} (51.76\% mIoU). 
While ESEG~\cite{ESEG2025AAAI}, which fine-tunes a SegFormer~\cite{xie2021segformer} decoder with edge supervision, maintains a slight lead in mIoU (57.55\%), \methodname attains comparable performance without requiring specialized edge-guidance modules or complex recurrent decoders. 
We report a detailed per-class IoU and the corresponding confusion matrix in the Appendix~\autoref{app:results}.

\subsection{Feature Matching}
\label{sec:feature_matching}
 \begin{table}[t]
\centering
\setlength{\tabcolsep}{4pt} %
    \begin{tabular}{l|c|ccccc}
        \toprule
        \textbf{Dataset} & \textbf{Metric} & \textbf{LLAK} & \textbf{RATE} & \textbf{EventPoint} & \textbf{SuperEvent} & \textbf{\methodname} \\
        & \textbf{AUC} & \cite{chiberre2022llak} & \cite{ikura2024rate} & \cite{huang2023wacv} & \cite{Burkhardt2025ICCV} &(Ours) \\
        \midrule
        \multirow{3}{*}{ECD} & $\mathbf{@5^{\circ}}$ $\uparrow$ & 0.7 & 3.3 & 1.6 & \ul{22.7} & \textbf{26.2} \\ %
        & \colortab $\mathbf{@10^{\circ}}$ $\uparrow$ & \colortab 1.4 & \colortab 8.4 & \colortab 3.0 & \colortab \ul{35.8} & \colortab \textbf{46.8} \\ %
        & $\mathbf{@20^{\circ}}$ $\uparrow$ & 2.1 & 18.0 & 5.4 & \ul{46.7} & \textbf{63.3} \\ %
        \midrule
        \multirow{3}{*}{EDS} & $\mathbf{@5^{\circ}}$ $\uparrow$ & 0.5 & 2.1 & 1.6 & \ul{15.2} &  \textbf{18.3} \\
        & \colortab $\mathbf{@10^{\circ}}$ $\uparrow$ & \colortab 0.7 & \colortab 5.1 & \colortab 2.8 & \colortab \ul{26.4} & \colortab \textbf{34.1} \\
        & $\mathbf{@20^{\circ}}$ $\uparrow$ & 1.0 & 10.3 & 5.2 & \ul{40.1} & \textbf{55.3} \\
        \bottomrule
    \end{tabular}
\vspace{2mm}
\caption{\textbf{Quantitative results on the ECD~\cite{mueggler2017event} and EDS~\cite{hidalgo2022cvpr} datasets.} We report the Area Under Curve (AUC) in \% at different thresholds.}
\label{tab:pose_estimation_eval}
\vspace*{-5mm}
\end{table}

\begin{table}[t]
\centering
\footnotesize
\setlength{\tabcolsep}{3.9pt}
\begin{tabular}{l|l|ccc}
\toprule
\textbf{Modality} & \textbf{Metric} & \textbf{MINIMA}~\cite{ren2025minima} & \textbf{SuperEvent}~\cite{Burkhardt2025ICCV} & \textbf{\methodname} (Ours)\\
\midrule

\multirow{5}{*}{Event-Event}
& AUC@5° $\uparrow$  & \ul{39.0} & 5.7 & \textbf{47.7} \\
& \colortab AUC@10° $\uparrow$ & \colortab \ul{53.9} & \colortab 11.4 & \colortab \textbf{64.9} \\
& AUC@20° $\uparrow$ & \ul{68.0} & 19.6 & \textbf{79.1} \\
& \colortab Med. Err (°) $\downarrow$ & \colortab \ul{3.93} & \colortab 46.33 & \colortab \textbf{2.75} \\
& S. Rate (\%) $\uparrow$   & \textbf{100.0} & \ul{99.9} & 99.7 \\

\midrule

\multirow{5}{*}{Image-Event}
& AUC@5° $\uparrow$  & 23.8 & - & \textbf{26.0} \\
& \colortab AUC@10° $\uparrow$ & \colortab 39.4 & \colortab - & \colortab \textbf{46.3} \\
& AUC@20° $\uparrow$ & 55.1 & - & \textbf{64.6} \\
& \colortab Med. Err (°) $\downarrow$ & \colortab 6.46 & \colortab - & \colortab \textbf{5.45} \\
& S. Rate (\%) $\uparrow$   & \textbf{100.0} & - & 99.6 \\

\bottomrule
\end{tabular}
\vspace{2mm}
\caption{\textbf{Pose estimation performance grouped by sensing modality.} Higher AUC in \% and success rate are better, while a lower median angular error is better. The experiments are conducted on the VECtor~\cite{gao22ral} dataset.}
\label{tab:wideview_results}
\vspace*{-7mm}
\end{table}

We evaluate the quality of the learned latent space by benchmarking \methodname on wide-baseline feature matching and relative pose estimation. Unlike prior works that rely on task-specific descriptors, we equip \methodname with the MASt3R~\cite{Mast3r2025eccv} head. Crucially, this geometric head remains frozen and was trained exclusively on RGB data. This setup serves as the ultimate test for our cross-modal alignment: if \methodname correctly maps events into the DUNE manifold, the MASt3R~\cite{Mast3r2025eccv} head should treat event features as native RGB representations.

\vspace*{-0.8mm}%
\paragraph{Experimental Setup.} We evaluate on the ECD~\cite{mueggler2017event}, EDS~\cite{hidalgo2022cvpr}, and VECtor~\cite{gao22ral} datasets. Following the protocol in~\cite{huang2023wacv}, we report the Area Under Curve (AUC) for relative rotation error at thresholds of 5°, 10°, and 20°, as well as the median angular error. We compare against specialized event-based baselines, including LLAK~\cite{chiberre2022llak}, RATE~\cite{ikura2024rate}, EventPoint~\cite{huang2023wacv}, SuperEvent~\cite{Burkhardt2025ICCV}, and the recent cross-modal MINIMA~\cite{ren2025minima}.

\vspace*{-0.8mm}%
\paragraph{Quantitative Results.} As summarized in~\autoref{tab:pose_estimation_eval} and ~\autoref{fig:matches_visual}, \methodname outperforms all existing methods by a significant margin. On the ECD~\cite{mueggler2017event} dataset, our method achieves an AUC@20° of 63.3\%, representing a substantial improvement over the previous state-of-the-art SuperEvent~\cite{Burkhardt2025ICCV} (46.7\%). These results demonstrate that leveraging the geometric priors of an RGB foundation model is more effective than training specialized event descriptors from scratch. 

\vspace*{-0.8mm}%
\paragraph{Wide-Baseline Robustness.} To further probe the robustness of our alignment, we conduct a wide-viewpoint evaluation following the protocol in~\cite{polizzi2025WACV}. We match a query event stream against targets at increasing angular distances. 
\autoref{tab:wideview_results} shows that while MINIMA~\cite{ren2025minima} degrades as the viewpoint change increases, \methodname is capable of producing more accurate feature matches, leading to lower orientation errors. 
\methodname, leveraging the MASt3R~\cite{Mast3r2025eccv} decoder, which provides strong geometric cues, outperforms the other methods in both event-event and image-event matching tasks. 
We refer the reader to~\autoref{app:results} for further analysis.

Notably, \methodname achieves these results in a zero-shot manner, whereas MINIMA~\cite{ren2025minima} requires a specialized generative data engine and fine-tuning to reach generalization. This highlights that our joint LoRA-based optimization effectively "unlocks" universal geometric reasoning for the event modality.
We report in the Appendix a breakdown of per-scene evaluation in~\autoref{app:results}.

\section{Limitations and Future Work}
\label{sec:limitations}

Our approach has two main limitations that also point to directions for future work. First, our convolutional embedder operates on fixed-size voxel grids, limiting its flexibility across sensors of differing resolution; we currently
bridge this gap with padding and overlapping tiling (Secs.~\ref{sec:depth_estimation} and~\ref{sec:semantic_seg}). A resolution-agnostic embedder operating directly on the asynchronous stream would remove this constraint. Second, the embedder does not explicitly model long-term temporal dynamics. We handle low-event
periods with a memory-holding mechanism, which is effective but remains a practical workaround; asynchronous or recurrent embedders are a principled next step, maintaining temporal consistency through sparse periods without frame-level heuristics.

\section{Conclusion}
\label{sec:conclusion}

We have presented \methodname, a cross-modal visual encoder that leverages large-scale RGB pretraining to achieve state-of-the-art performance in event-based vision. 
By employing a cross-modality alignment strategy, \methodname effectively bridges the gap between asynchronous event streams and intensity frames, enabling the direct use of foundation models originally designed for image data. 
Our results demonstrate that \methodname successfully inherits the deep geometric and semantic priors of the DUNE~\cite{Sariyildiz2025CVPR} teacher. Most notably, this alignment enables the direct, zero-shot application of complex image-trained decoders, such as MASt3R~\cite{Mast3r2025eccv}, to raw event data—outperforming specialized architectures.
Finally, by showing that event data can inhabit the same manifold as RGB foundation models, \methodname provides a solution to the scarcity of annotated event datasets and unlocks a broad ecosystem of RGB-based techniques, from calibration-free SLAM to cross-modal localization, for the event domain.

\section*{Acknowledgments}
This work was supported by the Natural Sciences and Engineering Research Council of Canada (NSERC) through the RGPIN program and the Canada Research Chairs program, the Canada Foundation for Innovation, and the Ontario Research Fund. This research was enabled in part by support provided by Compute Ontario and the Digital Research Alliance of Canada.

{
    \small
    \bibliographystyle{splncs04}
    \bibliography{full} 

@STRING{aaai	= "Proc. {AAAI} Conf. Artif. Intell." }

@STRING{arxiv	= "ar{X}iv e-prints" }

@STRING{cvpr	= "Proc. IEEE Conf. Comput. Vis. Pattern Recognit. (CVPR)" }

@STRING{cvprw	= "Proc. IEEE Conf. Comput. Vis. Pattern Recognit. Workshops (CVPRW)" }

@STRING{eccv	= "Eur. Conf. Comput. Vis. (ECCV)" }

@STRING{iccv	= "Int. Conf. Comput. Vis. (ICCV)" }

@STRING{ijcv	= "Int. J. Comput. Vis." }

@STRING{iros	= "IEEE/RSJ Int. Conf. Intell. Robot. Syst. (IROS)" }

@STRING{pami	= "{IEEE} Trans. Pattern Anal. Mach. Intell." }

@STRING{sciro	= "Sci. Robot." }

@STRING{tmlr	= "Trans. Mach. Learn. Res." }

@STRING{tvcg	= {IEEE Trans. Vis. Comput. Graph.} }

@STRING{wacv	= "{IEEE/CVF} Winter Conf. Appl. Comput. Vis. (WACV)" }

@STRING{tcsvt   = "{IEEE} Trans. Circuits Syst. Video Technol."}

@STRING{bmvc    = "Proc. Br. Mach. Vis. Conf. (BMVC)"}

@STRING{iclr    = "Int. Conf. Learn. Represent. (ICLR)"}

@STRING{threedv = "Int. Conf. 3D Vis. (3DV)"}

@STRING{ral = "{IEEE} Robot. Autom. Lett."}

@Article{	  botao2024microsaccade,
   author = {He, Botao and Wang, Ze and Zhou, Yuan and Chen, Jingxi and Singh, Chahat Deep and Li, Haojia and Gao, Yuman and Shen, Shaojie and Wang, Kaiwei and Cao, Yanjun and Xu, Chao and Aloimonos, Yiannis and Gao, Fei and Fermüller, Cornelia},
  journal	= sciro,
  title		= {Microsaccade-Inspired Event Camera for Robotics},
  year		= {2024}
}

@Article{	  gallego2022tpami,
  author	= {Gallego, Guillermo and Delbrück, Tobi and Orchard,
		  Garrick and Bartolozzi, Chiara and Taba, Brian and Censi,
		  Andrea and Leutenegger, Stefan and Davison, Andrew J. and
		  Conradt, Jörg and Daniilidis, Kostas and Scaramuzza,
		  Davide},
  journal	= pami,
  title		= {Event-Based Vision: A Survey},
  year		= {2022}
}

@Article{	  gehrig2021dsec,
  author	= {Gehrig, Mathias and Aarents, Willem and Gehrig, Daniel and
		  Scaramuzza, Davide},
  journal	= ral,
  title		= {{DSEC}: A Stereo Event Camera Dataset for Driving Scenarios},
  year		= {2021}
}

@Article{	  gehrig2024low,
  author	= {Gehrig, Daniel and Scaramuzza, Davide},
  journal	= {Nature},
  title		= {Low-latency automotive vision with event cameras},
  year		= {2024}
}

@InProceedings{	  kong2024openess,
  author	= {Kong, Lingdong and Liu, Youquan and Ng, Lai Xing and Cottereau, Benoit R. and Ooi, Wei Tsang},
  booktitle	= cvpr,
  title		= {{OpenESS}: Event-Based Semantic Scene Understanding with Open Vocabularies},
  year		= {2024}
}

@InProceedings{	  messikommer2023data,
  author	= {Messikommer, Nico and Fang, Carter and Gehrig, Mathias and
		  Scaramuzza, Davide},
  booktitle	= cvpr,
  title		= {Data-Driven Feature Tracking for Event Cameras},
  year		= {2023}
}

@Article{	  mueggler2017event,
  author	= {Mueggler, Elias and Rebecq, Henri and Gallego, Guillermo
		  and Delbrück, Tobi and Scaramuzza, Davide},
  journal	= {Int. J. Rob. Res.},
  title		= {The Event-Camera Dataset and Simulator: Event-based Data for Pose Estimation, Visual Odometry, and {SLAM}},
  year		= {2017}
}

@Article{	  rebecq19pami,
   author={Rebecq, Henri and Ranftl, René and Koltun, Vladlen and Scaramuzza, Davide},
  journal=pami, 
  title={High Speed and High Dynamic Range Video with an Event Camera}, 
  year={2021},
  volume={43},
  number={6},
  pages={1964-1980}
}

@article{rosinol_2018_ral,
   author={Vidal, Antoni Rosinol and Rebecq, Henri and Horstschaefer, Timo and Scaramuzza, Davide},
  journal=ral, 
  title={Ultimate {SLAM}? Combining Events, Images, and {IMU} for Robust Visual {SLAM} in {HDR} and High-Speed Scenarios}, 
  year={2018},
  volume={3},
  number={2},
  pages={994-1001}
}

@Article{	  shariff2024event,
  author={Shariff, Waseem and Dilmaghani, Mehdi Sefidgar and Kielty, Paul and Moustafa, Mohamed and Lemley, Joe and Corcoran, Peter},
  journal={IEEE Access}, 
  title={Event Cameras in Automotive Sensing: A Review}, 
  year={2024},
  volume={12},
  pages={51275-51306}
}

@Article{	  oquab2024dinov2,
  author	= {Oquab, Maxime and Darcet, Timothée and Moutakanni, Theo
		  and Vo, Huy V. and Szafraniec, Marc and Khalidov, Vasil and
		  Fernandez, Pierre and Haziza, Daniel and Massa, Francisco
		  and El-Nouby, Alaaeldin and Howes, Russell and Huang,
		  Po-Yao and Xu, Hu and Sharma, Vasu and Li, Shang-Wen and
		  Galuba, Wojciech and Rabbat, Mike and Assran, Mido and
		  Ballas, Nicolas and Synnaeve, Gabriel and Misra, Ishan and
		  Jegou, Herve and Mairal, Julien and Labatut, Patrick and
		  Joulin, Armand and Bojanowski, Piotr},
  journal	= tmlr,
  title		= {{DINOv2}: {Learning} Robust Visual Features without
		  Supervision},
  year		= {2024}
}

@InProceedings{	  caron2021dino,
  title		= {Emerging Properties in Self-Supervised Vision
		  Transformers},
  author	= {Caron, Mathilde and Touvron, Hugo and Misra, Ishan and
		  J\'egou, Herv\'e and Mairal, Julien and Bojanowski, Piotr
		  and Joulin, Armand},
  booktitle	= iccv,
  year		= {2021}
}

@Article{simeoni2025dinov3,
  title		= {{DINOv3}},
  author	= {Sim{\'e}oni, Oriane and Vo, Huy V. and Seitzer, Maximilian
		  and Baldassarre, Federico and Oquab, Maxime and Jose, Cijo
		  and Khalidov, Vasil and Szafraniec, Marc and Yi, Seungeun
		  and Ramamonjisoa, Micha{\"e}l and Massa, Francisco and
		  Haziza, Daniel and Wehrstedt, Luca and Wang, Jianyuan and
		  Darcet, Timoth{\'e}e and Moutakanni, Th{\'e}o and Sentana,
		  Leonel and Roberts, Claire and Vedaldi, Andrea and Tolan,
		  Jamie and Brandt, John and Couprie, Camille and Mairal,
		  Julien and J{\'e}gou, Herv{\'e} and Labatut, Patrick and
		  Bojanowski, Piotr},
  year		= {2025},
  eprint  = {2508.10104},
  journal	= arxiv
}

@INPROCEEDINGS{polizzi2025vibes,
  author	= {Polizzi, Vincenzo and Yang, Stephen and Clark, Quentin and Kelly, Jonathan and Gilitschenski, Igor and Lindell, David B.},
  booktitle	= threedv, 
  title	 	= {{VibES}: Induced Vibration for Persistent Event-Based Sensing}, 
  year		= {2026}
}

@InProceedings{	  zhu2019cvpr,
  author={Zhu, Alex Zihao and Yuan, Liangzhe and Chaney, Kenneth and Daniilidis, Kostas},
  booktitle=cvpr, 
  title={Unsupervised Event-Based Learning of Optical Flow, Depth, and Egomotion}, 
  year={2019},
  pages={989-997},
}

@InProceedings{	  sariyildiz2025cvpr,
  author	= {Sar{\i}y{\i}ld{\i}z, Mert B\"ulent and Weinzaepfel,
		  Philippe and Lucas, Thomas and de Jorge, Pau and Larlus,
		  Diane and Kalantidis, Yannis},
  title		= {{DUNE}: Distilling a Universal Encoder from Heterogeneous {2D} and {3D} Teachers},
  booktitle	= cvpr,
  month		= {June},
  year		= {2025},
  pages		= {30084-30094}
}

@inproceedings{rebecq2017bmvc,
  title={Real-time Visual-Inertial Odometry for Event Cameras using Keyframe-based Nonlinear Optimization.},
  author={Rebecq, Henri and Horstschaefer, Timo and Scaramuzza, Davide},
  booktitle=bmvc,
  year={2017}
}

@InProceedings{	  gehrig2019iccv,
  author	= {Gehrig, Daniel and Loquercio, Antonio and Derpanis,
		  Konstantinos G. and Scaramuzza, Davide},
  title		= {End-to-End Learning of Representations for Asynchronous
		  Event-Based Data},
  booktitle	= iccv,
  month		= {October},
  year		= {2019}
}

@inproceedings{hu2022lora,
  title     = {{LoRA}: Low-Rank Adaptation of Large Language Models},
  author    = {Hu, Edward J. and Shen, Yelong and Wallis, Phillip and Allen-Zhu, Zeyuan and Li, Yuanzhi and Wang, Shean and Wang, Lu and Chen, Weizhu},
  booktitle = iclr,
  year      = {2022}
}

@InProceedings{	  baradel2024mhr,
  title		= {{Multi-HMR}: Multi-person Whole-Body Human Mesh Recovery in a Single Shot},
  author	= {Baradel, Fabien and Armando, Matthieu and Galaaoui, Salma
		  and Br{\'e}gier, Romain and Weinzaepfel, Philippe and
		  Rogez, Gr{\'e}gory and Lucas, Thomas},
  booktitle	= eccv,
  pages		= {202--218},
  year		= {2024},
  organization	= {Springer}
}

@Article{	  badrinarayanan2017segnet,
  author={Badrinarayanan, Vijay and Kendall, Alex and Cipolla, Roberto},
  journal=pami, 
  title={{SegNet}: A Deep Convolutional Encoder-Decoder Architecture for Image Segmentation}, 
  year={2017},
  volume={39},
  number={12},
  pages={2481-2495},
}

@InProceedings{	  alonso2019cvpr,
  author={Alonso, I{\~n}igo and Murillo, Ana C.},
  booktitle=cvprw, 
  title={{EV-SegNet}: Semantic Segmentation for Event-Based Cameras}, 
  year={2019},
  pages={1624-1633},
}

@InProceedings{	  wang2021cvpr,
  author	= {Wang, Lin and Chae, Yujeong and Yoon, Sung-Hoon and Kim,
		  Tae-Kyun and Yoon, Kuk-Jin},
  title		= {{EvDistill}: Asynchronous Events To End-Task Learning via
		  Bidirectional Reconstruction-Guided Cross-Modal Knowledge
		  Distillation},
  booktitle	= cvpr,
  month		= {June},
  year		= {2021},
  pages		= {608-619}
}

@InProceedings{	  zhu2023iros,
   author={Zhu, Junyu and Liu, Lina and Jiang, Bofeng and Wen, Feng and Zhang, Hongbo and Li, Wanlong and Liu, Yong},
  booktitle=iros, 
  title={Self-Supervised Event-Based Monocular Depth Estimation Using Cross-Modal Consistency}, 
  year={2023},
  pages={7704-7710},
}

@InProceedings{	  hidalgo20203dv,
author={Hidalgo-Carrió, Javier and Gehrig, Daniel and Scaramuzza, Davide},
  booktitle=threedv, 
  title={Learning Monocular Dense Depth from Events}, 
  year={2020},
  pages={534-542}
}

@InProceedings{	  bhat2021cvpr,
  title={{AdaBins}: Depth Estimation Using Adaptive Bins},
  author={Bhat, Shariq Farooq and Alhashim, Ibraheem and Wonka, Peter},
  booktitle	= cvpr,
  pages		= {4009--4018},
  year		= {2021}
}

@InProceedings{	  yuan2022cvpr,  
title={Neural Window Fully-connected {CRFs} for Monocular Depth Estimation}, 
  author	= {Yuan, Weihao and Gu, Xiaodong and Dai, Zuozhuo and Zhu,
		  Siyu and Tan, Ping},
  booktitle	= cvpr,
  pages		= {3916--3925},
  year		= {2022}
}

@inproceedings{Mast3r2025eccv,
  title={Grounding Image Matching in {3D} with {MASt3R}},
  author={Leroy, Vincent and Cabon, Yohann and Revaud, J{\'e}r{\^o}me},
  booktitle=eccv,
  pages={71--91},
  year={2024},
  organization={Springer}
}

@Article{	  gehrig2021ral,
  title		= {Combining Events and Frames Using Recurrent Asynchronous Multimodal Networks for Monocular Depth Prediction},
  author	= {Gehrig, Daniel and R{\"u}egg, Michelle and Gehrig, Mathias
		  and Hidalgo-Carri{\'o}, Javier and Scaramuzza, Davide},
  journal	= ral,
  volume	= {6},
  number	= {2},
  pages		= {2822--2829},
  year		= {2021}
}

@Article{	  lowe2004sift,
    title={Distinctive Image Features from Scale-Invariant Keypoints}, 
  author	= {Lowe, David G.},
  journal	= ijcv,
  volume	= {60},
  number	= {2},
  pages		= {91--110},
  year		= {2004},
  publisher	= {Springer}
}

@InProceedings{	  rublee2011orb,
  title		= {{ORB}: An efficient alternative to {SIFT} or {SURF}},
  author	= {Rublee, Ethan and Rabaud, Vincent and Konolige, Kurt and
		  Bradski, Gary},
  booktitle	= iccv,
  pages		= {2564--2571},
  year		= {2011},
  organization	= {IEEE}
}

@inproceedings{	  bay2008surf,
  title		= {{SURF}: Speeded Up Robust Features},
  author	= {Bay, Herbert and Tuytelaars, Tinne and Van Gool, Luc},
  booktitle	= eccv,
  pages   = {404--417},
  year    = {2006}
}

@InProceedings{	  leutenegger2011brisk,
  title		= {{BRISK}: Binary robust invariant scalable keypoints},
  author	= {Leutenegger, Stefan and Chli, Margarita and Siegwart,
		  Roland Y},
  booktitle	= iccv,
  pages		= {2548--2555},
  year		= {2011},
  organization	= {IEEE}
}

@InProceedings{	  detone2018superpoint,
  title		= {{SuperPoint}: Self-supervised interest point detection and
		  description},
  author	= {DeTone, Daniel and Malisiewicz, Tomasz and Rabinovich,
		  Andrew},
  booktitle	= cvprw,
  pages		= {224--236},
  year		= {2018}
}

@inproceedings{sarlin20cvpr,
  title={{SuperGlue}: Learning feature matching with graph neural networks},
  author={Sarlin, Paul-Edouard and DeTone, Daniel and Malisiewicz, Tomasz and Rabinovich, Andrew},
  booktitle=cvpr,
  pages={4938--4947},
  year={2020}
}

@InProceedings{	  yi16eccv,
  author	= {Yi, Kwang Moo and Trulls, Eduard and Lepetit, Vincent and Fua, Pascal},
  title		= {{LIFT}: Learned invariant feature transform},
  booktitle	= eccv,
  year		= 2016,
  pages		= {467--483}
}

@InProceedings{	  wang2025vggt,
  title		= {{VGGT}: Visual Geometry Grounded Transformer},
  author	= {Wang, Jianyuan and Chen, Minghao and Karaev, Nikita and
		  Vedaldi, Andrea and Rupprecht, Christian and Novotny,
		  David},
  booktitle	= cvpr,
  year		= {2025}
}

@InProceedings{	  dust3rcvpr24,
  title		= {{DUSt3R}: Geometric {3D} Vision Made Easy},
  author	= {Wang, Shuzhe and Leroy, Vincent and Cabon, Yohann and Chidlovskii, Boris and Revaud, Jerome},
  booktitle	= cvpr,
  year		= {2024}
}

@InProceedings{	  hidalgo2022cvpr,
  title		= {Event-aided direct sparse odometry},
  author	= {Hidalgo-Carri{\'o}, Javier and Gallego, Guillermo and
		  Scaramuzza, Davide},
  booktitle	= cvpr,
  pages		= {5781--5790},
  year		= {2022}
}

@Article{	  zhu2018mvsec,
  title		= {The Multivehicle Stereo Event Camera Dataset: An Event Camera Dataset for {3D} Perception},
  author	= {Zhu, Alex Zihao and Thakur, Dinesh and {\"O}zaslan, Tolga and Pfrommer, Bernd and Kumar, Vijay and Daniilidis, Kostas},
  journal	= ral,
  volume	= {3},
  number	= {3},
  pages		= {2032--2039},
  year		= {2018}
}

@Article{	  gao22ral,
  author	= {Gao, Ling and Liang, Yuxuan and Yang, Jiaqi and Wu,
		  Shaoxun and Wang, Chenyu and Chen, Jiaben and Kneip,
		  Laurent},
  journal	= ral,
  title		= {{VECtor}: A Versatile Event-Centric Benchmark for
		  Multi-Sensor {SLAM}},
  year		= {2022},
  volume	= {7},
  number	= {3},
  pages		= {8217-8224}
}

@InProceedings{	  gehrig3dv2021,
  author	= {Gehrig, Mathias and Millh{\"a}usler, Mario and Gehrig, Daniel and Scaramuzza, Davide},
  title		= {{E-RAFT}: Dense Optical Flow from Event Cameras},
  booktitle	= threedv,
  year		= {2021},
  pages 	= {197-206}
}

@InProceedings{	  chaney2023cvpr,
  author={Chaney, Kenneth and Cladera, Fernando and Wang, Ziyun and Bisulco, Anthony and Hsieh, M. Ani and Korpela, Christopher and Kumar, Vijay and Taylor, Camillo J. and Daniilidis, Kostas},
  booktitle=cvprw, 
  title={{M3ED}: Multi-Robot, Multi-Sensor, Multi-Environment Event Dataset}, 
  year={2023},
  pages={4016-4023},
}

@Article{	  Hori2025TVCG,
  author	= {Hori, Ryosuke and Isogawa, Mariko and Mikami, Dan and
		  Saito, Hideo},
  journal	= tvcg,
  title		= {{EventPointMesh}: Human Mesh Recovery Solely From Event
		  Point Clouds},
  year		= {2025},
  issn		= {1941-0506},
  pages		= {5593-5610},
  volume  = {31},
  number  = {9},
  month		= sep
}

@inproceedings{	  eseg2025aaai,
  title		= {{ESEG}: Event-Based Segmentation Boosted by Explicit Edge-Semantic Guidance},
  booktitle	= aaai,
  author	= {Zhao, Yucheng and Lyu, Gengyu and Li, Ke and Wang, Zihao and Chen, Hao and Yang, Zhen and Deng, Yongjian},
  volume={39},
  pages={10510--10518},
  year={2025}
}

@Article{	  chiberre2022llak,
  title		= {Long-lived accurate keypoints in event streams},
  author	= {Chiberre, Philippe and Perot, Etienne and Sironi, Amos and Lepetit, Vincent},
  journal	= arxiv,
  eprint={2209.10385},
  year		= {2022}
}

@inproceedings{	  ikura2024rate,
  title		= {{RATE}: Real-time asynchronous feature tracking with event cameras},
  author	= {Ikura, Mikihiro and Le Gentil, Cedric and M{\"u}ller,
		  Marcus G and Schuler, Florian and Yamashita, Atsushi and
		  St{\"u}rzl, Wolfgang},
  booktitle	= iros,
  pages		= {11662--11669},
  year		= {2024}
}

@InProceedings{	  huang2023wacv,
  title		= {{EventPoint}: Self-supervised interest point detection and
		  description for event-based camera},
  author	= {Huang, Ze and Sun, Li and Zhao, Cheng and Li, Song and Su,
		  Songzhi},
  booktitle	= wacv,
  pages		= {5396--5405},
  year		= {2023}
}

@inproceedings{	  burkhardt2025iccv,
  title={{SuperEvent}: Cross-modal learning of event-based keypoint detection for {SLAM}},
  author={Burkhardt, Yannick and Schaefer, Simon and Leutenegger, Stefan},
  booktitle=iccv,
  pages={8918--8928},
  year={2025}
}

@inproceedings{	  sun2022ess,
  title		= {{ESS}: Learning Event-based Semantic Segmentation from Still Images},
  author	= {Sun, Zhaoning and Messikommer, Nico and Gehrig, Daniel and Scaramuzza, Davide},
  booktitle	= eccv,
  pages		= {341--357},
  year		= {2022},
  organization	= {Springer}
}

@InProceedings{	  cheng2022masked,
  title		= {Masked-attention mask transformer for universal image segmentation},
  author	= {Cheng, Bowen and Misra, Ishan and Schwing, Alexander G and Kirillov, Alexander and Girdhar, Rohit},
  booktitle	= cvpr,
  pages		= {1290--1299},
  year		= {2022}
}

@Article{	  xie2021segformer,
  title		= {{SegFormer}: Simple and efficient design for semantic
		  segmentation with transformers},
  author	= {Xie, Enze and Wang, Wenhai and Yu, Zhiding and Anandkumar,
		  Anima and Alvarez, Jose M and Luo, Ping},
  journal	= {Adv. Neural Inf. Process. Syst.},
  volume	= {34},
  pages		= {12077--12090},
  year		= {2021}
}

@InProceedings{	  ren2025minima,
  title		= {{MINIMA}: Modality Invariant Image Matching},
  author	= {Ren, Jiangwei and Jiang, Xingyu and Li, Zizhuo and Liang,
		  Dingkang and Zhou, Xin and Bai, Xiang},
  booktitle	= cvpr,
  year		= {2025}
}

@InProceedings{	  polizzi2025wacv,
  author	= {Polizzi, Vincenzo and Cannici, Marco and Scaramuzza,
		  Davide and Kelly, Jonathan},
  title		= {{FaVoR}: Features via Voxel Rendering for Camera
		  Relocalization},
  booktitle	= wacv,
  month		= {February},
  year		= {2025},
  pages		= {44-53}
}

@InProceedings{	  das2024halsie,
  title		= {{HALSIE}: Hybrid approach to learning segmentation by
		  simultaneously exploiting image and event modalities},
  author	= {Das Biswas, Shristi and Kosta, Adarsh and Liyanagedera,
		  Chamika and Apolinario, Marco and Roy, Kaushik},
  booktitle	= wacv,
  pages		= {5964--5974},
  year		= {2024}
}

@InProceedings{	  zubic2023iccv,
  author	= {Zubi\'c, Nikola and Gehrig, Daniel and Gehrig, Mathias and
		  Scaramuzza, Davide},
  title		= {From Chaos Comes Order: Ordering Event Representations for
		  Object Recognition and Detection},
  booktitle	= iccv,
  month		= {October},
  year		= {2023},
  pages		= {12846-12856}
}

@InProceedings{	  deng2022cvpr,
  title		= {A voxel graph {CNN} for object classification with event
		  cameras},
  author	= {Deng, Yongjian and Chen, Hao and Liu, Hai and Li, Youfu},
  booktitle	= cvpr,
  pages		= {1172--1181},
  year		= {2022}
}

@InProceedings{	  shiba22eccv,
  author	= {Shintaro Shiba and Yoshimitsu Aoki and Guillermo Gallego},
  title		= {Secrets of Event-based Optical Flow},
  booktitle	= eccv,
  pages		= {628--645},
  year		= 2022
}

@InProceedings{	  yang2024depthanything,
  title		= {{Depth Anything}: Unleashing the Power of Large-Scale
		  Unlabeled Data},
  author	= {Yang, Lihe and Kang, Bingyi and Huang, Zilong and Xu,
		  Xiaogang and Feng, Jiashi and Zhao, Hengshuang},
  booktitle	= cvpr,
  year		= {2024}
}

@Article{	  ghosh2025tpami,
  author	= {Ghosh, Suman and Gallego, Guillermo},
  journal	= pami,
  title		= {Event-Based Stereo Depth Estimation: A Survey},
  year		= {2025},
  volume	= {47},
  number	= {10},
  pages		= {9130-9149}
}

@InProceedings{kirillov2023segment,
    author    = {Kirillov, Alexander and Mintun, Eric and Ravi, Nikhila and Mao, Hanzi and Rolland, Chloe and Gustafson, Laura and Xiao, Tete and Whitehead, Spencer and Berg, Alexander C. and Lo, Wan-Yen and Dollar, Piotr and Girshick, Ross},
    title     = {Segment Anything},
    booktitle = iccv,
    month     = {October},
    year      = {2023},
    pages     = {4015-4026}
}

@Article{	  thisanke2023semantic,
  title		= {Semantic segmentation using vision transformers: A
		  survey},
  author	= {Thisanke, Hans and Deshan, Chamli and Chamith, Kavindu and
		  Seneviratne, Sachith and Vidanaarachchi, Rajith and Herath,
		  Damayanthi},
  journal	= {Eng. Appl. Artif. Intell.},
  volume	= {126},
  pages		= {106669},
  year		= {2023},
  publisher	= {Elsevier}
}

@InProceedings{	  he2022mae,
  title		= {Masked autoencoders are scalable vision learners},
  author	= {He, Kaiming and Chen, Xinlei and Xie, Saining and Li,
		  Yanghao and Doll{\'a}r, Piotr and Girshick, Ross},
  booktitle	= cvpr,
  pages		= {16000--16009},
  year		= {2022}
}

@Article{	  mcinnes2018umap,
  year		= {2018},
  volume	= {3},
  number	= {29},
  pages		= {861},
  author	= {McInnes, Leland and Healy, John and Saul, Nathaniel and
		  Großberger, Lukas},
  title		= {{UMAP}: Uniform Manifold Approximation and Projection},
  journal	= {J. Open Source Softw.}
}

@Article{	  eigen2014depth,
  title		= {Depth Map Prediction From a Single Image Using A
		  Multi-Scale Deep Network},
  author	= {Eigen, David and Puhrsch, Christian and Fergus, Rob},
  journal	= {Adv. Neural Inf. Process. Syst.},
  volume	= {27},
  year		= {2014}
}

@inproceedings{zhu2025iccv,
  title		= {{Depth Any Event Stream}: Enhancing Event-based Monocular Depth Estimation via Dense-to-Sparse Distillation},
  author	= {Zhu, Jinjing and Pan, Tianbo and Cao, Zidong and Liu, Yexin and Kwok, James T and Xiong, Hui},
  booktitle	= iccv,
  pages		= {5146--5155},
  year		= {2025}
}

@article{liu2024ereformer,
 author={Liu, Xu and Li, Jianing and Shi, Jinqiao and Fan, Xiaopeng and Tian, Yonghong and Zhao, Debin},
  journal=tcsvt, 
  title={Event-Based Monocular Depth Estimation With Recurrent Transformers}, 
  year={2024},
  volume={34},
  number={8},
  pages={7417-7429}
}

@InProceedings{bartolomei2025iccv,
    author={Bartolomei, Luca and Mannocci, Enrico and Tosi, Fabio and Poggi, Matteo and Mattoccia, Stefano},
  booktitle=iccv, 
  title={Depth {AnyEvent}: A Cross-Modal Distillation Paradigm for Event-Based Monocular Depth Estimation}, 
  year={2025},
  pages={19669-19678}
}

@INPROCEEDINGS{li2018megadepth,
  author={Li, Zhengqi and Snavely, Noah},
  booktitle=cvpr, 
  title={{MegaDepth}: Learning Single-View Depth Prediction from Internet Photos}, 
  year={2018},
  pages={2041-2050},
}

@inproceedings{silberman2012eccv,
  title     = {Indoor Segmentation and Support Inference from {RGBD} Images},
  author    = {Silberman, Nathan and Hoiem, Derek and Kohli, Pushmeet and Fergus, Rob},
  booktitle = eccv,
  pages     = {746--760},
  year      = {2012}
}

@InProceedings{	  sudre2017dice,
  title		= {Generalised dice overlap as a deep learning loss function
		  for highly unbalanced segmentations},
  author	= {Sudre, Carole H and Li, Wenqi and Vercauteren, Tom and
		  Ourselin, Sebastien and Jorge Cardoso, M},
  booktitle	= {Proc. Int. Workshop Deep Learn. Med. Image Anal.},
  pages		= {240--248},
  year		= {2017},
  organization	= {Springer}
}

@InProceedings{	  lin2017iccv,
  title		= {Focal Loss for Dense Object Detection},
  author	= {Lin, Tsung-Yi and Goyal, Priya and Girshick, Ross and He,
		  Kaiming and Doll{\'a}r, Piotr},
  booktitle	= iccv,
  pages		= {2980--2988},
  year		= {2017}
}
}


\clearpage

\appendix
\renewcommand{\sectionautorefname}{App.}
\setcounter{page}{1}
\setcounter{section}{0}
\setcounter{figure}{0}
\setcounter{table}{0}
\setcounter{equation}{0}
\renewcommand{\thesection}{S\arabic{section}}      %
\renewcommand{\theHsection}{appendix.\thesection}   %
\renewcommand{\thefigure}{S\arabic{figure}}
\renewcommand{\thetable}{S\arabic{table}}
\renewcommand{\theequation}{S\arabic{equation}}

\begin{figure}[t!]
    \begin{center}
       \textbf{\Large Supplementary Material}\\
       \vspace{1em}
       \textbf{\large REALM: An RGB- and Event-Aligned\\[4pt] Latent Manifold for Cross-Modal Perception}\\
       \vspace{1em}
       Vincenzo Polizzi\textsuperscript{1} \quad David B. Lindell\textsuperscript{2} \quad Jonathan Kelly\textsuperscript{1}\\
       \vspace{0.5em}
       \textsuperscript{1}University of Toronto, Robotics Institute\\
       \textsuperscript{2}University of Toronto, Department of Computer Science
    \end{center}
    \vspace{1em}
\end{figure}

This document provides additional technical details, architectural specifications, and extended experimental results that complement our manuscript. For additional visual results, we refer the reader to the accompanying video.

\section{Training Details}
\label{app:training}
\subsection{LoRA Rank Analysis}
\begin{table}[t]
\centering
\setlength{\tabcolsep}{6pt}
\begin{tabular}{cl|c|cc|c}
\toprule
\textbf{Rank} & \textbf{LoRA} & \textbf{Depth} & \multicolumn{2}{c|}{\textbf{Segmentation}} & \textbf{Loss} \\
 & \textbf{Modules} & \textbf{RMSE\,$\downarrow$} & \textbf{mIoU\,$\uparrow$} & \textbf{Acc\,$\uparrow$} & $\downarrow$ \\
\midrule
\multirow{2}{*}{32} & Att              & 10.26 & 10.06 & 48.83 & 0.172 \\
                    & \colortab Att+FFW & \colortab 10.55 & \colortab \tb{11.69} & \colortab \tb{51.12} & \colortab \tb{0.149} \\
\midrule
16 & Att+FFW & 10.48 & 11.18 & 49.70 & 0.156 \\
\midrule
8  & \colortab Att+FFW & \colortab \tb{10.25} & \colortab 10.82 & \colortab 49.35 & \colortab 0.164 \\
\bottomrule
\end{tabular}
\vspace{3mm}
\caption{\textbf{Effect of LoRA rank on downstream task performance.}
We apply LoRA either to the attention layers (\textit{Att}) or to both attention and feed-forward layers (\textit{Att+FFW}). Trained only using the EPM dataset~\cite{Hori2025TVCG}.}
\label{tab:lora_study}
\vspace*{-5mm}
\end{table}

\autoref{tab:lora_study} reports the effect of the LoRA rank on downstream performance. We evaluate three ranks \{8, 16, 32\}, corresponding to 0.7M, 2.4M, and 4.8M trainable parameters, respectively, on semantic segmentation and depth estimation. Rank 32 with Att+FFW achieves the best segmentation and lowest loss.  Although these values are comparable to the rank 16 evaluation, we adopt a LoRA rank of 32 and an alpha of 64 as the default for our evaluation.

\subsection{Masking Effects}
\begin{figure}[t!]
\centering
  \centering
  \includegraphics[width=\linewidth]{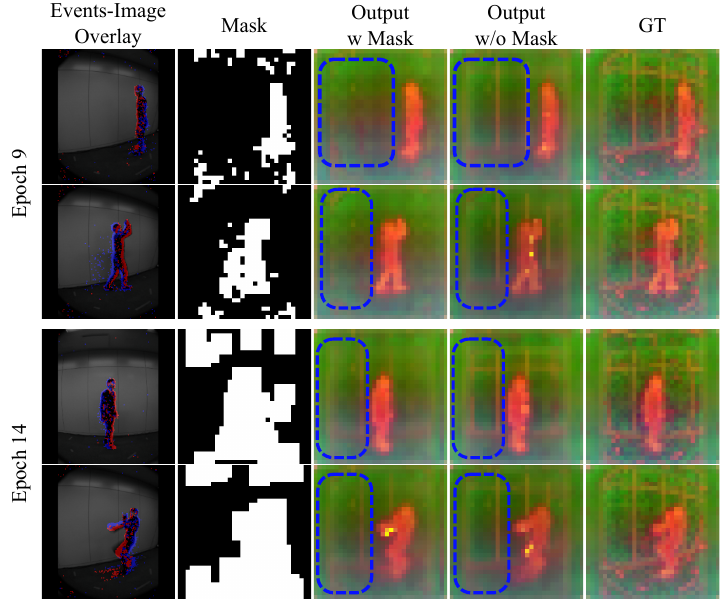}
\caption{\textbf{Qualitative evolution of the masking strategy during distillation.} We illustrate the impact of progressive masking on the \methodname feature space. The student is fed with events (see Events-Image Overlay), and the resulting features are visualized via PCA against the ground truth (GT) from the RGB-based DUNE~\cite{Sariyildiz2025CVPR} teacher. Without the mask (w/o Mask), the student overfits to the teacher in regions lacking event data (e.g., notice the spurious background lines highlighted in blue). Applying the mask (w/ Mask) focuses the learning strictly on regions with active events. This prevents the model from overfitting to unobserved parts of the scene, where predicting teacher features would require unfounded hallucinations merely to minimize the loss. By Epoch 14, the mask radius has dilated from 1 token (Epochs 1–9) to 2 tokens, where one token corresponds to a 14 $\times$ 14 pixel region. Consequently, the student at Epoch 14 successfully learns to extrapolate and "hallucinate" valid context further away from the active event edges, avoiding the severe overfitting seen in the unmasked baseline. All feature maps and masks are rescaled to the $448 \times 448$ input resolution for visualization.}
\label{fig:masking}
\end{figure}

\begin{table}[t]
\centering
\setlength{\tabcolsep}{6pt}
\begin{tabular}{c|c|cc|c}
\toprule
\textbf{Masks} & \textbf{Depth} & \multicolumn{2}{c|}{\textbf{Segmentation}} & \textbf{Loss} \\
 & \textbf{RMSE\,$\downarrow$} & \textbf{mIoU\,$\uparrow$} & \textbf{Acc\,$\uparrow$} & $\downarrow$ \\
\midrule
\nomark & 8.42 & 53.97 & 88.84 & 0.250 \\
\colorrow \yesmark & \tb{8.06} & \tb{55.47} & \tb{89.30} & \tb{0.246} \\
\bottomrule
\end{tabular}
\vspace{3mm}
\caption{\textbf{Effect of masking strategy on downstream task performance.}
We evaluate the effect of the masking on the downstream tasks of segmentation and depth estimation. Models are trained on the entire dataset cluster.}
\label{tab:mask_study}
\vspace*{-5mm}
\end{table}

In~\autoref{fig:masking}, we illustrate the effect of progressive spatial distillation during training. Our objective is to align the output of our event-based encoder with the target features extracted by DUNE~\cite{Sariyildiz2025CVPR}. Specifically, while DUNE generates representations from the RGB image, our model is trained to predict those exact features using the event stream from a time window before the image timestamp.

Notably, we aim to prevent \methodname from hallucinating or overfitting to static parts of the scene that are unobservable in the event stream. Because event cameras only capture dynamic changes, a static background visible in the RGB image will produce no events. Consequently, our model should not be forced to reconstruct features for these regions, because the model lacks the underlying sensory information to do so.

By applying a spatial mask to our distillation objective, we strictly avoid penalizing the model in regions where no events occur. As shown in~\autoref{fig:masking}, this strategy successfully prevents the model from forcing a representation of the static background. In contrast, training without this masking causes the model to completely overfit to the teacher, incorrectly generating features for unobserved areas.

While successfully preventing background overfitting, this masking strategy introduces a potential pitfall:
if the student model is penalized exclusively on event-active pixels, it tends to learn a trivial, sparse representation. Essentially, the model degrades into an edge detector, failing to infer dense object volumes. To mitigate this, we leverage the inherent contextual reasoning of our Masked Autoencoder (MAE)~\cite{he2022mae} architecture. While the MAE strategy forces the network to infer missing spatial information and fill in the gaps, its effectiveness is bottlenecked if the distillation loss only evaluates sparse edge pixels. Therefore, we complement the MAE objective with a progressive spatial masking schedule. By gradually dilating the distillation mask to expose the full scene for a limited number of epochs, we allow \methodname to fully utilize its MAE-driven contextual understanding without overfitting to unobserved static regions. For the dilation schedule implementation details see~\autoref{app:method_training}.

To evaluate this progressive masking strategy, we utilize the EventPointMesh \cite{Hori2025TVCG} (EPM) dataset. 
EPM~\cite{Hori2025TVCG} features extensive static backgrounds, making it an ideal benchmark to validate our approach to avoid background overfitting. As demonstrated in~\autoref{fig:masking}, across various training epochs and dilation radii, \methodname successfully suppresses the hallucination of unobserved static regions while preserving highly accurate structural representations of moving objects, i.e., the human subjects.

We report in~\autoref{tab:mask_study} the effect of the proposed masking strategy on downstream performance. We train \methodname with a LoRA rank of 16, with and without masking, and evaluate on semantic segmentation and depth estimation. The masked variant outperforms the unmasked baseline across both tasks, improving mIoU and accuracy on segmentation and reducing RMSE on depth, while also achieving a lower distillation loss. This confirms that progressive spatial masking prevents the student from overfitting to unobserved static regions and yields a more transferable latent space.
\subsection{\methodname Training}
\label{app:method_training}
We train the model for 30 epochs with an effective batch size of 512 (128 per GPU across 4 accumulation steps) and a gradient clipping norm of 1.0. 
Optimization is performed using AdamW with a base learning rate of $1 \times 10^{-3}$ and a weight decay of 0.01. The student model is fine-tuned using LoRA with a rank $r=32$ and $\alpha=64$. We apply a 10\% dropout to the adapted attention and feed-forward modules ($qkv$, $proj$, $fc1$, $fc2$). 
While the base \methodname architecture contains 91,096,000 parameters, we only optimize the voxel embedding layer (5,232,832 params) and the injected LoRA adapters (4,803,072 params). 
This results in a total of 10,035,904 trainable parameters, representing approximately 10.47\% of the total model capacity (see~\autoref{tab:model_parameters}).

The distillation loss penalizes differences between
 the teacher and student's normalized CLS and patch tokens (\verb|x_norm_clstoken|, \verb|x_norm_patchtokens|) using a composite loss function, training also the normalization layer of DUNE~\cite{Sariyildiz2025CVPR}. The loss is weighted with
$\lambda_{\text{MSE}}=0.1$, $\lambda_{\text{cos}}=0.3$, and $\lambda_{\text{L1}}=0.6$.

To ensure robust feature learning and force contextual hallucination, we introduce input token dropping (MAE~\cite{he2022mae}) with a 30\% probability starting at epoch 8.
Independently, for the spatial distillation loss, we employ a progressive curriculum. The spatial activity mask is strictly enforced initially to focus entirely on event edges.
We then progressively expand the mask's boundary outward into the background. Specifically, at epochs 10, 15, and 20, we morphologically dilate the initial base mask by a radius of 2, 4, and 6 patches, respectively (where each patch represents a $14 \times 14$ pixel region).

This base spatial mask is generated dynamically within the dataloader. We collapse the input event voxel grid across the temporal and channel dimensions to obtain a 2D spatial activity map. 
We then apply $14 \times 14$ max-pooling to downsample this map to match the spatial resolution of the vision transformer's patch tokens (forming a $32 \times 32$ grid for the $448 \times 448$ input). 
This yields a binary mask where a value of 1 indicates the presence of at least one event within that patch.
To dilate this mask during the curriculum phases, we reshape the flattened token mask back into its 2D grid format and apply a morphological dilation operation. 
Specifically, we utilize a 2D max-pooling operation with a $3 \times 3$ kernel, a stride of 1, and a padding of 1. A single iteration of this operation effectively expands the active mask regions outward by exactly one patch in all directions.

\section{Datasets Details}
\label{app:datasets}
\begin{table}[t]
\centering
\begin{tabular}{c|cc|ccccc}
\toprule
\textbf{Dataset} & \textbf{Indoor} & \textbf{Outdoor} & \textbf{Seg.} & \textbf{Depth} & \textbf{Pose} & \textbf{HMR} & \textbf{Syn.} \\
\midrule
DSEC~\cite{gehrig2021dsec, Gehrig3dv2021, kong2024openess}  & \nomark  & \yesmark   & \yesmark  & \yesmark  & \nomark    & \nomark  & \nomark\\
\colorrow EventScape~\cite{gehrig2021RAL}                   & \nomark  & \yesmark   & \yesmark  & \yesmark  & \nomark    & \nomark  & \yesmark \\
M3ED~\cite{chaney2023cvpr}                        & \nomark  & \yesmark   & \yesmark  & \yesmark  & \yesmark   & \nomark  & \nomark\\
\colorrow EDS~\cite{hidalgo2022cvpr}                        & \yesmark & \nomark    & \nomark   & \nomark   & \yesmark   & \nomark  & \nomark\\
EventPointMesh~\cite{Hori2025TVCG}                          & \yesmark & \nomark    & \nomark   & \nomark   & \nomark    & \yesmark & \nomark\\
\bottomrule
\end{tabular}
\vspace{3mm}
\caption{\textbf{Multimodal datasets used for distillation.} We report the diverse collection of event-RGB datasets utilized to train the \methodname embedding and LoRA layers. These datasets are selected to cover a wide array of scenarios, including indoor navigation, outdoor driving, and human-centric motion to provide the network with a comprehensive overview of diverse geometric and semantic structures.}
\label{tab:datasets}
\vspace*{-5mm}
\end{table}

To make the training resilient to varying event densities (which depend on the motion present in the scene), we employ a fixed time window for certain datasets and a minimum amount of events for others (see below for more details). Given the diversity of the datasets used (see~\autoref{tab:datasets}), we expose the model to different dynamics and event throughputs.

From the datasets, we manually select training scenes that minimize motion blur and extreme darkness to ensure high-quality distillation from the DUNE~\cite{Sariyildiz2025CVPR} teacher.

\paragraph{Datasets Preparation.} Training requires temporally and spatially synchronized event-RGB pairs. We segment the event stream into windows bounded by consecutive RGB frames, using either fixed time intervals or event-count thresholds to ensure resilience across diverse scene dynamics and throughputs. 

To achieve spatial alignment, we warp the RGB frames into the event camera's coordinate system using known intrinsic and extrinsic parameters. This avoids artifacts in the event representation and ensures both modalities have the same geometric resolution. 

Notably, we process raw, distorted event data rather than attempting to undistort them. This preserves the model's robustness to varying camera intrinsics, a property consistent with the capabilities of its teacher backbone. This rigorous preparation is necessitated by the scarcity of complete multimodal event datasets, as many existing benchmarks lack the depth or synchronization required for precise cross-modal alignment.

\paragraph{Dataset Parameters.}To ensure consistent feature extraction, all events are converted into a voxel grid representation with five temporal bins and normalized before being processed. 
We adopt a target spatial resolution of $448 \times 448$ across all modalities, applying center cropping and resizing to maintain geometric consistency with the teacher model.
To handle diverse scene dynamics, we employ a hybrid windowing strategy:
\begin{itemize}
    \item \textbf{Fixed-Count Windows:} For DSEC and EventScape, we use a threshold of 150,000 events to ensure sufficient signal density in high-motion or complex environments.
    \item \textbf{Fixed-Time Windows:} For EDS, M3ED, and EPM, we segment the stream into 33 ms intervals, corresponding to standard video frame rates, to maintain temporal alignment with the teacher’s RGB input.
\end{itemize}
The spatial alignment pipeline utilizes known intrinsic and extrinsic parameters to warp RGB frames for EDS and M3ED. Additionally, we apply a stride to the sequences (a stride of 2 for DSEC, EventScape, and M3ED, and 5 for EPM) to maximize scene diversity during training.

\section{Output Head Training}
\label{app:heads_training}

\subsection{Segmentation Head}
To train the segmentation head, we used the images provided in the DSEC~\cite{gehrig2021dsec} dataset with the corresponding semantic labels. We used 11 class labels, as in the ESS~\cite{sun2022ess} work.

We train the head on the frozen DUNE~\cite{Sariyildiz2025CVPR} backbone and then use it zero-shot on the \methodname backbone. The number of parameters for the segmentation head is reported in~\autoref{tab:model_parameters}.
The training lasts for 300 epochs with a batch size of 128 and a gradient clipping maximum norm of 1.0.
Optimization is performed using the AdamW optimizer with a learning rate of $1 \times 10^{-4}$ and a weight decay of 0.01.
To align features for the task, we attach to the backbone the DINOv2 projector as suggested in DUNE~\cite{Sariyildiz2025CVPR} when performing semantic segmentation. 
The training objective is a composite loss function combining the multi-class Focal loss~\cite{lin2017iccv} and the Dice loss~\cite{sudre2017dice}. Formally, the total loss is defined as:
\begin{align}
\mathcal{L}_{\text{total}} = \lambda_{\text{Dice}} \mathcal{L}_{\text{Dice}} + \lambda_{\text{Focal}} \mathcal{L}_{\text{Focal}}
\end{align}
To explicitly address the severe class imbalance inherent in driving scenes—where background classes like roads and skies dominate the pixel count while small objects are heavily underrepresented—we apply a static class-weighting vector $\mathbf{w}$ to both loss formulations. 
For our 11 semantic classes (sky, building, fence, person, pole, road, sidewalk, vegetation, car, wall, traffic light), the weight vector is empirically defined to heavily penalize errors on minority classes, $\mathbf{w} = [1.0, 1.0, 5.0, 5.0, 5.0, 1.0, 3.0, 2.0, 2.5, 10.0, 15.0]$.

\subsection{Depth Head}
Similarly to the segmentation head, we use the images provided by the synthetic dataset DENSE~\cite{hidalgo20203dv} and the real data from MVSEC~\cite{zhu2018mvsec}, training on the frozen DUNE~\cite{Sariyildiz2025CVPR} backbone.

We follow a training procedure similar to E2Depth~\cite{hidalgo20203dv}, utilizing a two-stage curriculum: an initial pre-training phase of 150 epochs on the synthetic DENSE~\cite{hidalgo20203dv} dataset, followed by 150 epochs of fine-tuning on real data from MVSEC~\cite{zhu2018mvsec}. With the depth head, we do not use any projector from DUNE~\cite{Sariyildiz2025CVPR}.

Optimization is performed with an effective batch size of 128 and a gradient clipping maximum norm of 1.0. 
During training, we mask out invalid, infinite, or missing depth values. Ground-truth depth targets are strictly clamped to a valid range of $[1.95, 82.0]$ meters to stabilize early training and prevent supervision from sensor noise at extreme distances.
The depth training objective is a composite loss function operating in the logarithmic space, combining a Scale-Invariant (SI) log loss~\cite{eigen2014depth} and a Multi-Scale Gradient (MSG) loss~\cite{li2018megadepth}. 
The total depth loss is formulated 
as:
\begin{align}
\mathcal{L}_{\text{total}} = \lambda_{\text{SI}} \mathcal{L}_{\text{SI}} + \lambda_{\text{MSG}} \mathcal{L}_{\text{MSG}}
\end{align}
where we empirically set the weighting coefficients to $\lambda_{\text{SI}} = 2.0$ and $\lambda_{\text{MSG}} = 0.01$.
To penalize relative depth errors while remaining robust to global scale ambiguities, we apply the Scale-Invariant log loss. Let $d_i = \log(\hat{y}_i) - \log(y_i)$ denote the logarithmic difference between the predicted depth $\hat{y}_i$ and the ground-truth depth $y_i$ at valid pixel $i$. 

Optimization is performed using the AdamW optimizer with a refined learning rate of $5 \times 10^{-5}$ and a weight decay of 0.01. 

\subsection{MASt3R Decoder}

We do not perform any training on the MASt3R~\cite{Mast3r2025eccv} decoder, and use it directly on \methodname output. In particular, we use the refined decoder provided in DUNE~\cite{Sariyildiz2025CVPR}.

\section{Results}
\label{app:results}

\subsection{Ablation Study on Events Representations}
\begin{table}[h]
\centering
\begin{tabular}{lcc|ccc}
\toprule
\textbf{Task} & \textbf{Dataset} & \textbf{Metrics} & \textbf{Voxel Grid} & \textbf{Tencode} & \textbf{ERGO} \\
\midrule
\multirow{2}{*}{Sem. Segmentation} & \multirow{2}{*}{DSEC~\cite{gehrig2021dsec}} & \textbf{IoU} $\uparrow$ [\%] & \textbf{53.75} & 50.64 & 50.62 \\ 
 &  & \colortab \textbf{Accuracy} $\uparrow$ [\%] & \colortab \textbf{88.95} & \colortab 87.86 & \colortab 87.87 \\ 
\midrule
\multirow{4}{*}{Feat. Matching} & \multirow{2}{*}{VECtor~\cite{gao22ral}} & \textbf{Med. Err.} $\downarrow$ [$^\circ$] & \textbf{10.00} & 10.63 & 11.55 \\ 
 &  & \colortab \textbf{AUC} $\mathbf{@5^{\circ}}$ $\uparrow$ & \colortab \textbf{0.016} & \colortab 0.013 & \colortab \textbf{0.016} \\ 
 & \multirow{2}{*}{(robot-fast)} & \textbf{AUC} $\mathbf{@10^{\circ}}$ $\uparrow$ & \textbf{0.156} & 0.136 & 0.131 \\ 
 &  & \colortab \textbf{AUC} $\mathbf{@20^{\circ}}$ $\uparrow$ & \colortab \textbf{0.444} & \colortab 0.397 & \colortab 0.374 \\ 
\midrule
Depth Estimation & DENSE~\cite{hidalgo20203dv} & \textbf{RMSE} [m] $\downarrow$ & 8.654 & 8.283 & \textbf{8.151} \\ 
\bottomrule
\end{tabular}
\vspace{3mm}
\caption{\textbf{Quantitative comparison of event representations across downstream tasks.} We evaluate voxel grids~\cite{zhu2019cvpr}, Tencode~\cite{huang2023wacv}, and ERGO~\cite{Zubic2023ICCV} on semantic segmentation, feature matching, and depth estimation. While depth estimation results are comparable across representations, voxel grids demonstrate superior performance in semantic segmentation and wide-baseline matching robustness.}
\label{tab:representation_study}
\vspace*{-5mm}
\end{table}

We utilize a voxel-grid representation for feeding the event data to our encoder. This choice was informed by an ablation study comparing three prominent event encoders: Tencode~\cite{huang2023wacv}, ERGO~\cite{Zubic2023ICCV}, and voxel grids~\cite{zhu2019cvpr}. Each candidate was trained on a representative subset of our data and evaluated across three distinct downstream tasks: semantic segmentation, feature matching, and depth estimation.

While the models achieved comparable training losses across all representations, their downstream utility diverged. As shown in~\autoref{tab:representation_study}, the voxel-grid representation consistently outperformed the alternatives in semantic segmentation and feature matching. Although ERGO~\cite{Zubic2023ICCV} showed a slight advantage in depth estimation, the superior performance of voxel grids in high-level perception and matching tasks makes them the most suitable choice for our proposed architecture.

\subsection{Semantic Segmentation}
\begin{table}[t]
\centering
\setlength{\tabcolsep}{4pt}
\resizebox{\textwidth}{!}{
    \begin{tabular}{l|ccccccccccc}
        \toprule
         & \multicolumn{11}{c}{\textbf{Classes}} \\
        \cmidrule(lr){2-12}
        \textbf{Method} & Sky & Building & Fence & Person & Pole & Road & Sidewalk & Vegetation & Car & Wall & Traffic Light \\
        \midrule
        DUNE & 94.95 & 85.80 & 36.03 & 52.99 & 29.92 & 95.19 & 73.20 & 87.30 & 86.92 & 45.37 & 56.41 \\
        \colorrow \textbf{\methodname} & 91.28 & 79.25 & 18.90 & 32.71 & 15.45 & 90.28 & 50.00 & 80.39 & 78.35 & 38.07 & 34.35 \\
        \bottomrule
    \end{tabular}
}
\vspace{3mm}
\caption{\textbf{Per-class quantitative results.} We report the Intersection over Union (IoU) in \% for all individual classes, comparing our proposed method against the DUNE baseline.}
\label{tab:per_class_iou_comparison}
\vspace*{-5mm}
\end{table}

\begin{figure}[t!]
\centering
  \centering
  \includegraphics[width=\linewidth]{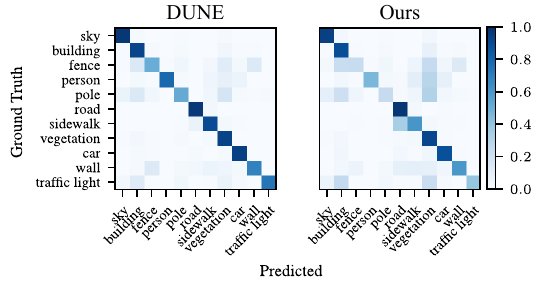}
\caption{\textbf{Confusion matrices for semantic segmentation.} We compare the predictions of the DUNE~\cite{Sariyildiz2025CVPR} teacher against our \methodname student across 11 classes. As expected, \methodname shows slightly more confusion—particularly on thin structures like poles and fences. Despite this, \methodname retains a highly pronounced diagonal, confirming that the student effectively captures the high-level semantic representations of the image-based teacher.}
\label{fig:conf_matrix}
\end{figure}

In~\autoref{tab:per_class_iou_comparison}, we report the per-class Intersection over Union (IoU) and~\autoref{fig:conf_matrix} shows the confusion matrix comparing DUNE~\cite{Sariyildiz2025CVPR} to \methodname.
We observe that compared to the image-based teacher DUNE~\cite{Sariyildiz2025CVPR}, our model exhibits lower performance on thin structures such as poles, fences, and traffic lights. 
This is likely attributed to the inherent sparsity of the event stream and the $14\times$ spatial downsampling in our patch embedding layer. 
However, the successful zero-shot application of the teacher's segmentation head to our student's features validates the high degree of alignment between the \methodname and DUNE~\cite{Sariyildiz2025CVPR} feature spaces.

\subsection{Matching with Wide Viewpoint Changes}
\begin{figure}[t!]
\centering
  \centering
  \includegraphics[width=\linewidth]{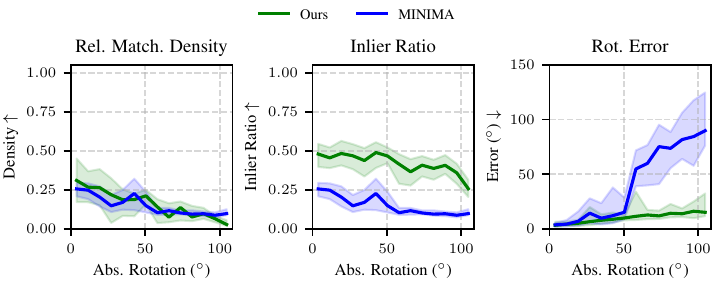}
  \includegraphics[width=\linewidth]{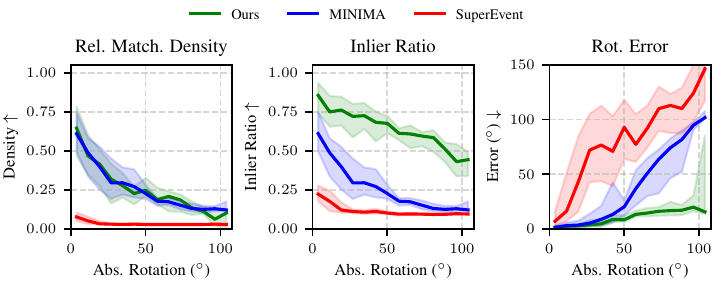}
\caption{\textbf{Robustness to wide-viewpoint changes on the VECtor~\cite{gao22ral} dataset.} The top row shows the evaluation of cross-modal (image–event) matching performance across increasing angular distances between query and target, and the second row shows the intra-modal matching (event-event). \textbf{(a) Relative Matching Density:} The ratio of RANSAC inliers to the maximum number of keypoints detected across the entire sequence.\textbf{(b) Inlier Ratio:} \methodname consistently maintains a higher percentage of correct correspondences as the rotation increases. \textbf{(c) Rotation Error:} Our method exhibits significantly higher stability and lower angular error compared to the specialized MINIMA~\cite{ren2025minima} baseline.}
\label{fig:wideview_eval}
\end{figure}

\begin{table}[ht]
\centering
\begin{tabular}{lc|cccc}
\toprule
\multicolumn{6}{c}{\textbf{Event--Event Matching}} \\
\midrule
\textbf{Scene} & \textbf{Model} & \textbf{0--15$^\circ$} & \textbf{15--30$^\circ$} & \textbf{30--45$^\circ$} & \boldmath{$> 45^\circ$} \\
& & (Easy) & (Med) & (Hard) & (Extreme) \\
\midrule
\multirow{3}{*}{hdr-fast} & SuperEvent~\cite{Burkhardt2025ICCV} & 0.029 & 0.004 & -- & --  \\ 
 & \colortab MINIMA~\cite{ren2025minima} & \colortab 0.527 & \colortab 0.346 & \colortab 0.138 & \colortab -- \\ 
 & \methodname & \textbf{0.789} & \textbf{0.687} & \textbf{0.460} & \textbf{0.088} \\ 
\midrule
\multirow{3}{*}{hdr-normal} & SuperEvent~\cite{Burkhardt2025ICCV} & 0.402 & 0.041 & 0.011 & -- \\ 
 & \colortab MINIMA~\cite{ren2025minima} & \colortab \textbf{0.910} & \colortab \textbf{0.726} & \colortab  0.605 & \colortab -- \\ 
 & \methodname & 0.880 & 0.656 & \textbf{0.858} & -- \\ 
\midrule
\multirow{3}{*}{robot-fast} & SuperEvent~\cite{Burkhardt2025ICCV} & 0.091 & 0.009 & -- & -- \\ 
 & \colortab MINIMA~\cite{ren2025minima} & \colortab 0.371 & \colortab 0.249 & \colortab 0.079 & \colortab 0.012 \\ 
 & \methodname & \textbf{0.635} & \textbf{0.545} & \textbf{0.282} & \textbf{0.139} \\ 
\midrule
\multirow{3}{*}{robot-normal} & SuperEvent~\cite{Burkhardt2025ICCV} & 0.269 & 0.231 & 0.052 & -- \\ 
 & \colortab MINIMA~\cite{ren2025minima} & \colortab 0.677 & \colortab 0.687 & \colortab \textbf{0.574} & \colortab 0.310 \\ 
 & \methodname & \textbf{0.796} & \textbf{0.741} & 0.320 & \textbf{0.342} \\ 
\bottomrule
\multicolumn{6}{c}{\textbf{Image--Event Matching}} \\
\midrule
\multirow{2}{*}{hdr-fast} & MINIMA~\cite{ren2025minima} & 0.316 & 0.154 & 0.128 & -- \\ 
 &  \colortab \methodname & \colortab \textbf{0.476} & \colortab \textbf{0.408} & \colortab \textbf{0.324} & \colortab \textbf{0.018} \\ 
\midrule
\multirow{2}{*}{hdr-normal} & MINIMA~\cite{ren2025minima} & 0.661 & 0.404 & \textbf{0.444} & -- \\ 
 & \colortab \methodname & \colortab \textbf{0.675} & \colortab \textbf{0.494} & \colortab 0.400 & \colortab -- \\ 
\midrule
\multirow{2}{*}{robot-fast} & MINIMA~\cite{ren2025minima} & 0.041 & 0.056 & 0.034 & 0.030 \\ 
 & \colortab \methodname & \colortab \textbf{0.288} & \colortab \textbf{0.219} & \colortab \textbf{0.162} & \colortab \textbf{0.115} \\ 
\midrule
\multirow{2}{*}{robot-normal} & MINIMA~\cite{ren2025minima} & \textbf{0.693} & \textbf{0.748} & \textbf{0.673} & 0.460 \\ 
 & \colortab \methodname & \colortab 0.618 & \colortab 0.657 & \colortab 0.484 & \colortab \textbf{0.463} \\ 
\bottomrule
\end{tabular}
\vspace{3mm}
\caption{Per-Scene Wide-Baseline Robustness. AUC at $10^\circ$ for various matching algorithms across different angular bins. We evaluate performance ranging from easy (0--$15^\circ$) to extreme ($> 45^\circ$) viewpoint variations. Our method, \methodname, consistently outperforms other methods in fast-motion scenarios across both event-event and image-event modalities. Notably, while MINIMA degrades significantly as viewpoint changes increase, \methodname maintains strong performance even under large rotations. This robustness to wide-baseline viewpoint changes stems directly from \methodname's use of the MASt3R decoder, which grounds feature matching in 3D geometry: by operating in a geometrically consistent latent space rather than relying on purely appearance-based descriptors, \methodname produces correspondences that are inherently more stable under large viewpoint changes. In normal-speed conditions, MINIMA~\cite{ren2025minima} achieves a marginally higher AUC at small baselines, though this difference is small and \methodname's advantage becomes clear and consistent as viewpoint changes increase beyond 30$^\circ$.}
\label{tab:per_scene_matches}
\vspace*{-4mm}
\end{table}

\autoref{fig:wideview_eval} evaluates image-event matching on the \textit{fast-robot} sequence of the VECtor~\cite{gao22ral} dataset. 
While MINIMA~\cite{ren2025minima} performance degrades significantly as viewpoint changes increase, \methodname maintains a higher inlier ratio and lower rotation error. 

The first row of \autoref{fig:wideview_eval} displays the relative matching density, defined as the ratio of successful \methodname inliers in a given pair relative to the maximum keypoints detected across the sequence.
Unlike the standard inlier ratio, this metric captures the absolute survival of feature correspondences under extreme viewpoint and modality shifts. 
Similarly, the second row illustrates event-event matching results, where our method consistently provides the highest inlier ratio and stable rotation estimation across varying baselines.

In~\autoref{tab:per_scene_matches}, we report a breakdown of the per-scene evaluation of the feature matching performance.

\subsection{Benchmark Feature Extraction}
\begin{table}[ht]
\centering
\begin{tabular}{l|ccc|cc}
\toprule
\textbf{Model} & \textbf{Extract} $\downarrow$ & \textbf{Match} $\downarrow$ & \textbf{Total} $\downarrow$ & \textbf{FPS} $\uparrow$ & \textbf{GPU} $\downarrow$ \\
 & \textbf{(ms)} & \textbf{(ms)} & \textbf{(ms)} &  & \textbf{(MB)} \\
\midrule
MINIMA & 167.2 $\pm$ 3.8 & 54.0 $\pm$ 0.1 & 221.2 $\pm$ 3.8 & 4.52 & 7563 \\
\colorrow \methodname (Ours) & \textbf{50.5} $\pm$ 0.1 & \textbf{60.8} $\pm$ 0.2 & \textbf{111.3} $\pm$ 0.2 & \textbf{8.99} & \textbf{2581} \\
\bottomrule
\end{tabular}
\vspace{3mm}
\caption{\textbf{Inference Performance Comparison.}}
\label{tab:inference_performance}
\vspace*{-5mm}
\end{table}

In~\autoref{tab:inference_performance} we evaluate the runtime performance of \methodname and MINIMA~\cite{ren2025minima} for the task of image-event matching. Our method runs with an inference time that is half of MINIMA~\cite{ren2025minima}.

Runtime performance is measured by running each model for five warm-up iterations, followed by 50 timed iterations on synthetic random inputs. Latency is broken down into two stages: feature extraction (the dense forward pass) and matching (keypoint sampling and correspondence estimation). 
We report mean, median, standard deviation, and latency in milliseconds, as well as throughput in FPS and peak GPU memory consumption. 
All models receive pre-processed tensors of identical spatial resolution before the timed block begins, so the preprocessing cost is excluded from all measurements equally.
By default, MINIMA~\cite{ren2025minima} operates at a coarse resolution of 560×560 followed by a high-resolution refinement pass at 864×864, regardless of input size. 
To ensure a fair comparison with the other models, we override MINIMA's internal resolution at construction time, effectively disabling the refinement stage. This aligns the computational budget across all models. 

\section{Models}
\label{app:models}
\begin{table}[t]
\centering
\setlength{\tabcolsep}{4pt}
\resizebox{\textwidth}{!}{
    \begin{tabular}{lcccc}
        \toprule
        \textbf{Component} &  \textbf{DUNE} & \textbf{REALM} & \textbf{Depth Head} & \textbf{Seg. Head} \\
         & \cite{Sariyildiz2025CVPR} & (Ours) & (Ours) & (Ours) \\
        \midrule
        Embedder  & 452,352 & 5,232,832 & --- & --- \\
        \colorrow Encoder & 85,863,168 & 85,863,168* & --- & --- \\
        \midrule
        \textbf{Backbone (Total)} & \textbf{86,315,520} & \textbf{91,096,000} & \textbf{91,096,000} & \textbf{91,096,000} \\
        \midrule
        \midrule
        \colorrow Projector & 0 & 0 & 0 & 7,874,560* \\
        Head & 0 & 0 & 393,728 & 11,275 \\
        \colorrow LoRA & 0 & 4,803,072 & 0 & 0 \\
        \midrule
        \textbf{Total Parameters} & \textbf{86,315,520} & \textbf{95,899,072} & \textbf{91,489,728} & \textbf{98,981,835} \\
        \bottomrule
    \end{tabular}
}
\vspace{2mm}
\caption{\textbf{Detailed Parameter Breakdown across Architecture Variants.} During REALM distillation training, only the voxel embedder and LoRA adapters are updated, all other parameters (DUNE backbone encoder, marked with *) remain frozen. During head training, the linear head weights are the only parameters updated; the backbone, the DINOv2 projector from DUNE (marked with *), and all other components remain frozen. This freezing policy is consistent across all training stages and at inference. The LoRA adapters are absorbed into the backbone at inference and are therefore not counted separately in the deployment footprint.}
\label{tab:model_parameters}
\vspace*{-5mm}
\end{table}

\begin{table}[ht]
\centering
\renewcommand{\arraystretch}{1.5} %
\begin{tabular}{l|ccc}
\toprule
 & \makecell{\textbf{Backbone} \\ \textbf{Stand Alone}} & \makecell{\textbf{Semantic} \\ \textbf{Segmentation}} & \makecell{\textbf{Depth} \\ \textbf{Estimation}} \\ 
\midrule
\textbf{Mean Time (us)} $\downarrow$ & 4416 $\pm$ 73 & 4660 $\pm$ 37 & 4649 $\pm$ 33 \\
\colorrow \textbf{GPU Peak Usage (MB)} $\downarrow$ & 435 & 464 & 436 \\
\textbf{Throughput (FPS)} $\uparrow$ & 226 & 215 & 215 \\
\bottomrule
\end{tabular}
\vspace{2mm}
\caption{\textbf{Runtime performance of the REALM backbone and task heads.} Mean inference time, peak GPU memory, and throughput measured over 1000 runs at 448×448 resolution with batch size 1 on a single GPU. The backbone accounts for the vast majority of both latency and memory cost. The marginal overhead introduced by the task heads is minimal: the depth head adds a single-channel output and costs under 1 MB of additional memory, while the segmentation head produces 11-channel logits over the full spatial resolution, accounting for the slightly higher peak memory usage of 464 MB.}
\label{tab:runtime_performance}
\vspace*{-5mm}
\end{table}

We describe the models utilized for the voxel embedding as well as the segmentation and depth heads. We report the number of parameters per model in~\autoref{tab:model_parameters}.

\subsection{Voxel Embedding}
The voxel embedding module is a specialized event stream architecture, designed to translate high-dimensional event data into the latent patch format required by the DUNE~\cite{Sariyildiz2025CVPR} transformer. It serves as the primary entry point for the \methodname model, replacing the lightweight RGB patch embedder with a convolutional encoder.

The module consists of a $7\times7$ convolutional stem followed by three progressive downsampling stages using residual encoder blocks. 
The resulting features are projected into a $768$-dimensional latent space and adaptively pooled to a $32\times32$ grid. 
This architecture allows the model to compress the sparse, temporal information of event voxel grids into dense patch tokens $(B, 1024, 768)$ that are structurally identical to the teacher's RGB embeddings.

\subsection{Segmentation Head}
For the semantic segmentation task, we utilize a head that performs a dense classification on the extracted features. 
The head consists of a $1 \times 1$ convolutional layer that maps the $768$-dimensional patch tokens to the $11$ target semantic classes. 
During the forward pass, the patch sequence is reconstructed into a spatial grid and projected to the class dimension. 
The resulting logits are then bilinearly upsampled to the original input resolution of $448 \times 448$ to produce the final segmentation mask.
The runtime evaluation for the semantic segmentation task is in~\autoref{tab:runtime_performance}.

\subsection{Depth Head}
The depth head implements a discrete binning strategy, also known as soft classification, to perform robust depth estimation. 
We define 256 depth bins linearly spaced between 1.95 m and 82.0 m. The architecture processes normalized features through two specialized paths:
\begin{itemize}
    \item \textbf{Spatial path:} A $1 \times 1$ convolutional layer projects the patch tokens to capture local spatial information, followed by an initial $4 \times$ bilinear interpolation.
    \item \textbf{Global path:} A linear layer projects the CLS token into the bin space, providing global scene context that is broadcast across the spatial dimensions.
\end{itemize}
The outputs of these paths are fused additively to produce a combined logit volume. 
A softmax activation is applied across the bin dimension to generate a probability distribution for each pixel. 
The final depth value is calculated as the expected value, allowing for differentiable training and sub-bin precision. 
The resulting depth map is then bilinearly upsampled to the target input resolution.
The runtime performance for the depth estimation task is reported in~\autoref{tab:runtime_performance}.

\section{Generalization to Other Tasks and Training on Different Datasets}
\label{app:extra_tasks}
\begin{figure}[t!]
\centering
  \centering
  \includegraphics[width=\linewidth]{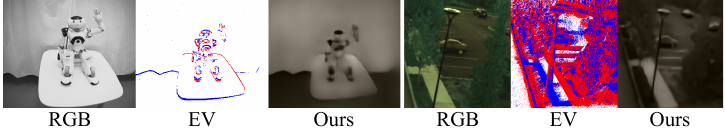}
\vspace{-7mm}
\caption{\textbf{Qualitative evaluation of image reconstruction.} The reconstruction head is trained using DUNE with data from the NYUv2~\cite{silberman2012eccv} and tested on events from the VECtor~\cite{gao22ral} and M3ED~\cite{chaney2023cvpr} datasets.}
\label{fig:img_reconstruction}
\vspace{-2ex}
\end{figure}

\begin{figure}[t!]
\centering
  \centering
  \includegraphics[width=\linewidth]{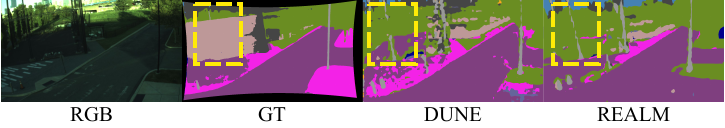}
\vspace{-7mm}
\caption{\textbf{Semantic segmentation on the M3ED~\cite{chaney2023cvpr} dataset.} The segmentation head is trained on the DSEC dataset and tested with M3ED~\cite{chaney2023cvpr} scenes.}
\label{fig:img_segmentation}
\vspace{-2ex}
\end{figure}

\begin{table}[ht]
\centering
\renewcommand{\arraystretch}{1.5}
\begin{tabular}{l|cc}
\toprule
\textbf{Task} & \textbf{DUNE} & \textbf{REALM} \\ \midrule
\textbf{Seg. (mIoU--Acc)} $\uparrow$ & 24.08 -- 65.71 & \tb{26.95} -- \tb{68.88} \\
\colorrow \textbf{Depth (RMSE)} $\downarrow$ & 3.25 & \tb{3.03} \\
\bottomrule
\end{tabular}
\vspace{2mm}
\caption{\tb{DUNE~\cite{Sariyildiz2025CVPR} (RGB) vs REALM (events) on M3ED~\cite{chaney2023cvpr} scenes.} We report the evaluation of DUNE~\cite{Sariyildiz2025CVPR} and REALM on scenes from the M3ED~\cite{chaney2023cvpr} dataset for the semantic segmentation and depth estimation tasks.}
\label{tab:dune_vs_realm}
\vspace*{-5mm}
\end{table}

In the main paper, we focus on depth and segmentation because they probe how geometrically and semantically grounded the latent space is.
Here, we show that the same aligned space generalizes more broadly, both to additional tasks and to training regimes that do not rely on event--RGB pairs from the target domain.

\subsection{Additional Tasks}
\autoref{fig:img_reconstruction} shows image reconstruction from event features using a 5M-parameter convolutional head trained on NYUv2~\cite{silberman2012eccv} ($\sim$47K images), evaluated on the \textit{fast-robot} sequence (VECtor) and the \textit{flight outdoor fast} sequence (M3ED). 
The same alignment that supports geometric and semantic heads thus also supports a generative decoder.

\subsection{Cross-Dataset Generalization}
A key property of REALM is that downstream heads need not be trained on event data, or even on the evaluation domain. 
To show this property, as well as the generalization capabilities of the latent space, we train our linear head on the NYUv2~\cite{silberman2012eccv} (an image-only dataset). Then we test on event-based datasets such as M3ED~\cite{chaney2023cvpr}. 
\autoref{fig:img_segmentation} shows that the mapping from the latent space remains consistent across datasets, allowing the model to generalize to unseen scenarios. 
M3ED's 1\,MP resolution is handled with the same overlapping-tile strategy used for DSEC~\cite{gehrig2021dsec} (\autoref{sec:semantic_seg}).

Under this full domain shift, \methodname surpasses its teacher DUNE on both depth and segmentation~\autoref{tab:dune_vs_realm}. 
\autoref{fig:img_segmentation} further shows REALM segmenting poles that are missed in the low-contrast RGB input and ground-truth labels, highlighting the value of event sensing: events can capture structure that standard RGB cameras miss under low contrast or adverse lighting.

\end{document}